\documentclass[runningheads]{llncs}
\usepackage[T1]{fontenc}
\usepackage{graphicx}

\usepackage{xcolor}
\definecolor{myred}{rgb}{.8,.0,.0}

\definecolor{myblue}{rgb}{.0,.0,.8}

\usepackage{url}
\usepackage[labelsep=space,labelfont=bf]{caption}
\usepackage{subcaption}
\usepackage{float}  
\usepackage{hyperref}
\usepackage{array, booktabs, multirow}
\usepackage{orcidlink}

\ifdefined\DOUBLEBLIND
\newcommand{\repourl}{\url{https://anonymous.4open.science/r/MaskOfTruth-D946} and \url{https://anonymous.4open.science/r/MaskOfTruth_EyeFundus-3FB2}}
\else
\newcommand{\repourl}{\url{https://github.com/TheoSourget/MMC_Masking} and \url{https://github.com/TheoSourget/MMC_Masking_EyeFundus}}
\fi
\begin{document}
\title{Mask of truth: model sensitivity to unexpected regions of medical images}

\ifdefined\DOUBLEBLIND
    \author{***}
    \authorrunning{***}
    \institute{***}
\else
    \author{Théo Sourget\inst{1}\orcidlink{0009-0005-0220-9590} \and
    Michelle Hestbek-Møller\inst{1}\orcidlink{0000-0001-6530-3078} \and
    Amelia Jiménez-Sánchez\inst{1}\orcidlink{0000-0001-7870-0603} \and \\
    Jack Junchi Xu\inst{2,3,4}\orcidlink{0000-0002-2259-6795} \and
    Veronika Cheplygina\inst{1}\orcidlink{0000-0003-0176-9324}}
    \institute{
     IT University of Copenhagen, Denmark \and
     Copenhagen University Hospital, Herlev and Gentofte, Denmark \and
     Radiological AI Testcenter, Denmark \and
     Department of Radiology, Zealand University Hospital, Denmark\\
    \email{\{tsou,vech\}@itu.dk}\\
    }
    \authorrunning{T. Sourget et al.}
\fi

\maketitle

\begin{abstract}
The development of larger models for medical image analysis has led to increased performance. However, it also affected our ability to explain and validate model decisions. Models can use non-relevant parts of images, also called spurious correlations or \emph{shortcuts}, to obtain high performance on benchmark datasets but fail in real-world scenarios. 
In this work, we challenge the capacity of convolutional neural networks (CNN) to classify chest X-rays and eye fundus images while masking out clinically relevant parts of the image. 
We show that all models trained on the PadChest dataset, irrespective of the masking strategy, are able to obtain an Area Under the Curve (AUC) above random. Moreover, the models trained on full images obtain good performance on images without the region of interest (ROI), even superior to the one obtained on images only containing the ROI. 
We also reveal a possible spurious correlation in the Chákṣu dataset while the performances are more aligned with the expectation of an unbiased model.
We go beyond the performance analysis with the usage of the explainability method SHAP and the analysis of embeddings. 
We asked a radiology resident to interpret chest X-rays under different masking to complement our findings with clinical knowledge.
Our code is available at \repourl


\vspace*{2ex}
\fbox{\parbox{0.90\linewidth}{This version of the article has been
accepted for publication, after peer review but is not the Version of Record and does not
reflect post-acceptance improvements, or any corrections. The Version of Record is available online at:
\url{http://dx.doi.org/10.1007/s10278-025-01531-5}.}}

\end{abstract}

\newpage
\section{Introduction}
\label{sec:intro}
With the increasing demand for imaging examinations and a shortage of clinicians to perform their analyses, the waiting time for patients has deteriorated. In this context, artificial intelligence (AI) has emerged as a possible solution and has become standard in certain applications as more devices received FDA approval~\cite{Wu2023ClinicalAdoption}.
This was possible thanks to the increased performances of AI models, sometimes even reported as outperforming domain experts on some aspects \cite{shen2019artificial,plesner2023autonomous}. 
However, this performance has been achieved using more complex models with less understanding of the reasons for the classification. This leads to biased models relying on non-relevant parts of the images or shortcuts correlated with the condition in the training data. It affects the evaluation of proposed methods resulting in an overestimation of their true capacities and poor performances when tested on external data~\cite{Wu2021DeviceEvaluation}.

While most research is focused on new methodologies improving performances, techniques to detect~\cite{jimenez2023detecting,oakden2020hidden} and mitigate~\cite{roschewitz2024counterfactual,boland2024all} these biases have been proposed. However, they often investigate a single type of bias and rely on prior knowledge or annotation of this bias.
Moreover, biases are often datasets specific, there is therefore a need to perform analyses on more recent datasets to discover their potential biases.
Finally, studies are generally limited to the effect on performance metrics, not assessing the representation of the data produced by models or including clinicians to evaluate their method. This last aspect seems essential to ensure the clinical relevancy of results when using generative models to create new images or masking out part of the image.

Our contributions, extending previous works on bias detection and aiming for a better understanding of the effect of spurious correlations, are as follows: 
\begin{enumerate}
    \item We evaluate the capacity of a CNN model to classify chest X-ray and eye fundus images with different masking strategies, highlighting the usage of non-clinically relevant parts of the images. 
    \item We perform a study with a radiology resident to evaluate the possibility of correctly classifying chest X-rays under different masking strategies.
    \item We use embedding analysis and an explainability method to better understand which elements are used but also discuss the limitations of such techniques in this setting.
    \item We discuss the limitations of our study and present possible extensions. 
\end{enumerate}

\section{Related Work}
Most studies on artificial intelligence and deep learning focus on presenting new methods to improve performances on benchmark datasets. In this context, CNNs have been widely applied to the medical field ~\cite{liz2023deep,de2023airogs,kassem2021machine}. To reduce the need for large training data of CNN models, capsule networks~\cite{sabour2017dynamic} that better model spatial relationships in images have also been applied to medical images~\cite{jimenez2018capsule,dos2020capsule}.
After the breakthrough of transformers on natural language processing tasks and their adaptation to computer vision, several works present approaches using transformer architectures with medical images models~\cite{shamshad2023transformers,marikkar2023lt,fan2023detecting} sometimes outperforming CNN models. To improve the generalization capacity of models, some studies present foundation models trained with large-scale datasets that can be applied to various downstream tasks~\cite{moutakanni2024advancing,wu2023towards}.
Finally, with the recent progress on models combining multiple types of data (e.g. image and text), some studies also assess vision-language models~\cite{liu2024visual,li2024visionunite}.
In our work, we do not focus on improving the classification metrics but rather on understanding model behaviours and the limitations of such evaluations. We therefore choose a widely used CNN classifier to conduct our analysis.

Recent studies show that datasets can suffer from problems like label errors, biases or shortcuts in datasets which may impact the reliability of the evaluation. In this work, we focus on shortcuts, also called spurious correlations or confounders, leading models to be biased and use irrelevant information to make decisions~\cite{geirhos2020shortcut,banerjee2023shortcuts,vasquez2024detecting}.

Some studies evaluate the impact of a specific shortcut like Jim\'{e}nez-S\'{a}nchez et al.~\cite{jimenez2023detecting} and Oakden-Rayner et al.~\cite{oakden2020hidden} who show the effect of chest drains on the classification of pneumothorax or Wargnier et al.~\cite{wargnier2024unexpected} who show the influence of the brain's shape in MRI classification. Boland et al.~\cite{boland2024there} also investigate which layers are impacted by spurious correlation looking at the difference between a model trained with a base non-biased dataset and a biased version of it.

Using generative models, other methods create new or synthetic data with or without the confounders for a better evaluation or to quantify the effect of specific confounders~\cite{weng2025fast,perez2024radedit}. 
Weng et al.~\cite{weng2025fast} use a diffusion model guided with an automatically generated mask of the area which should be modified to edit counterfactual images, detect spurious correlation and evaluate its impact. 
Pérez-García et al.~\cite{perez2024radedit} also use a diffusion model for data editing but indicate both which area should be modified and which area should be kept unchanged as well as a text prompt.

Some, more similar to our method, study the performance of models when removing the area of interest. Bissoto et al.~\cite{bissoto2019constructing} perform the analysis of melanoma classification removing the lesion in the image. Hemelings et al.~\cite{hemelings2021deep} also evaluate the performances of models on eye fundus photographies under different masking strategies showing good results despite removing the optic disc. Both studies only evaluate models on the same type of images (e.g. train and evaluate without the ROI) and did not assess whether clinicians were able to diagnose the images correctly. 
Haynes et al.~\cite{haynes2024generalisation} assess the capacity of Covid-19 classifiers to correctly classify images without the lungs. They find that models, especially when not pre-trained on related tasks, are still able to distinguish healthy and non-healthy images.
Aslani et al.~\cite{aslani2022optimising} present a pipeline improving COVID-19 detection in chest X-rays. The pipeline includes the segmentation of the lungs and crops the image to centralize the lungs, removing some possible biases on the image's borders such as texts.
Beyond classification models, shortcut learning is also a phenomenon impacting segmentation models as highlighted by Lin et al.~\cite{lin2024shortcut}.

Prior works showed the variety of possible shortcuts, Juodelyte et al.~\cite{juodelyte2024source} introduce a taxonomy for confounders and evaluate the robustness of CNN pre-trained on ImageNet~\cite{deng2009imagenet} and RadImageNet~\cite{mei2022radimagenet} for chest X-ray and CT datasets. They show that despite resulting in similar performances, models pre-trained on ImageNet are more impacted by shortcuts than models pre-trained on RadImageNet.

Being able to explain the outcome of a deep-learning model, often seen as a black box, is crucial, especially in a medical context. To this aim, explainability methods are applied to show the clinical relevancy of the image area used by models to classify samples \cite{liz2023deep,sun2023right}. Following the combination of image and text to improve performances, it is also used to analyse the data and model behaviours. Kim et al.~\cite{kim2024transparent} presents a framework with a foundation model to extract concepts in skin lesion images. These concepts can then be used to detect potential biases or be used as the input of a classification network to produce more explainable decisions following the idea of concept bottleneck models~\cite{koh2020concept}.

Our study adds to this line of research by analysing more recent datasets in two applications and including a domain expert to strengthen our findings with clinical knowledge. Moreover, our method doesn't rely on prior annotation of potential shortcuts but on the knowledge of clinically relevant parts of the image.

\section{Materials and Methods}
\subsection{Datasets}
\subsubsection{Chest X-rays:}
Our first task is chest X-ray classification using the PadChest dataset~\cite{Bustos2020PadChest}\footnote{Accessed from \url{https://bimcv.cipf.es/bimcv-projects/padchest/}}. We chose this dataset as it is less studied than more popular datasets such as CheXpert~\cite{Irvin2019Chexpert} or ChestX-ray14~\cite{Wang2017chestxray8}.
The dataset contains more than 160,000 images. We removed images of lateral views (marked as "L" in the "Projection" field of the metadata) and images for which the label was null or included "suboptimal study", "exclude" or "unchanged".

\begin{figure}[H]
    \centering
    \begin{subfigure}{.45\columnwidth}
      \centering
      \includegraphics[width=\linewidth]{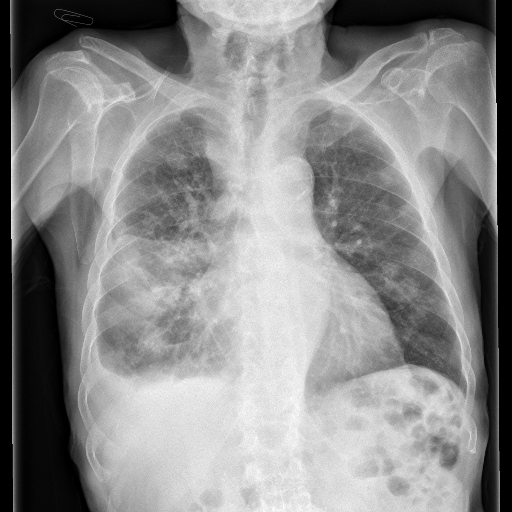}
        \caption{Chest X-ray from PadChest}
    \end{subfigure}
    \begin{subfigure}{.45\columnwidth}
        \centering
         \includegraphics[width=\linewidth]{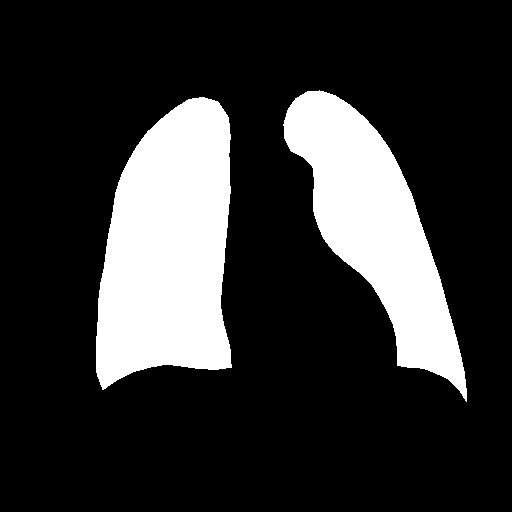}
      \caption{Associated mask from CheXmask}     
    \end{subfigure}
    \caption{Example of chest X-ray from PadChest dataset and associated mask from CheXmask}
    \label{fig:data_padchest_chexmask}
\end{figure}

The PadChest dataset doesn't provide segmentation masks for the lungs, we therefore decided to use the CheXmask dataset~\cite{gaggion2024chexmask}\footnote{Downloaded from \url{https://physionet.org/content/chexmask-cxr-segmentation-data/0.4/}}. The dataset contains the segmentation masks of five public datasets including PadChest created with a HybridGNet model and an automated quality assessment of the masks using the Reverse Classification Accuracy (RCA)~\cite{valindria2017reverse}. We only used masks for which the "Dice RCA (Mean)" is above 0.7 as advised by the dataset providers and we removed images for which no masks were available. \textbf{93,203 images were kept after this filtering}, examples of the data from PadChest and CheXmask can be seen in Fig.~\ref{fig:data_padchest_chexmask}. 
We also evaluate out-of-distribution (OOD) performance on 22,004 images from the ChestX-ray14 dataset's test set~\cite{Wang2017chestxray8}\footnote{Downloaded from \url{https://nihcc.app.box.com/v/ChestXray-NIHCC}}.

PadChest is a multi-class and multi-label dataset containing annotations for 174 labels. However, we decided to restrict our study to five classes also present in the ChestX-ray14 dataset: cardiomegaly, pneumonia, atelectasis, pneumothorax and effusion.

\subsubsection{Retinal fundus image:} 
\begin{figure}[H]
    \centering
    \begin{subfigure}{.3\columnwidth}
      \centering
      \includegraphics[width=\linewidth]{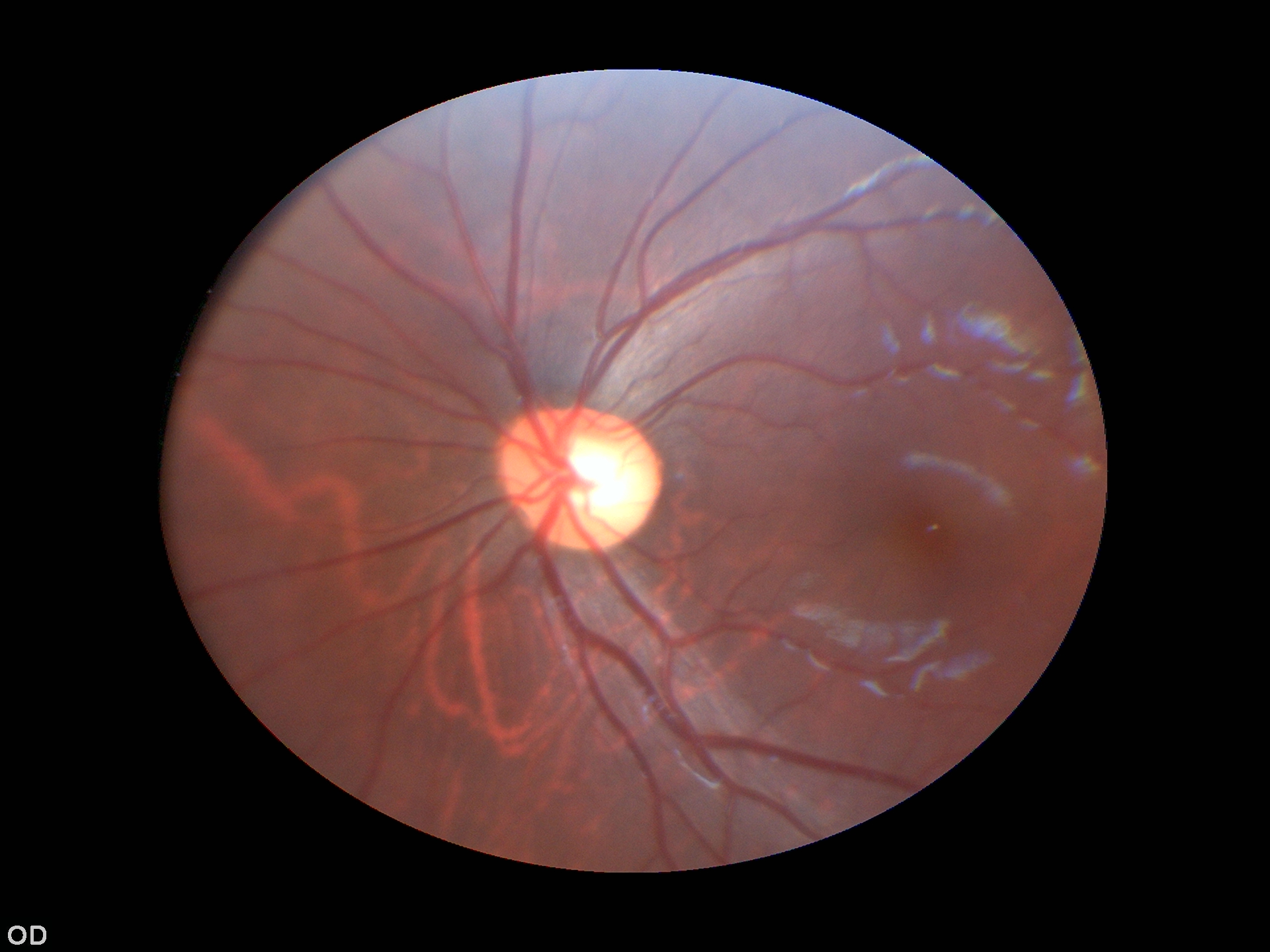}
        \caption{With Bosch camera}
    \end{subfigure}
    \begin{subfigure}{.3\columnwidth}
        \centering
         \includegraphics[width=\linewidth]{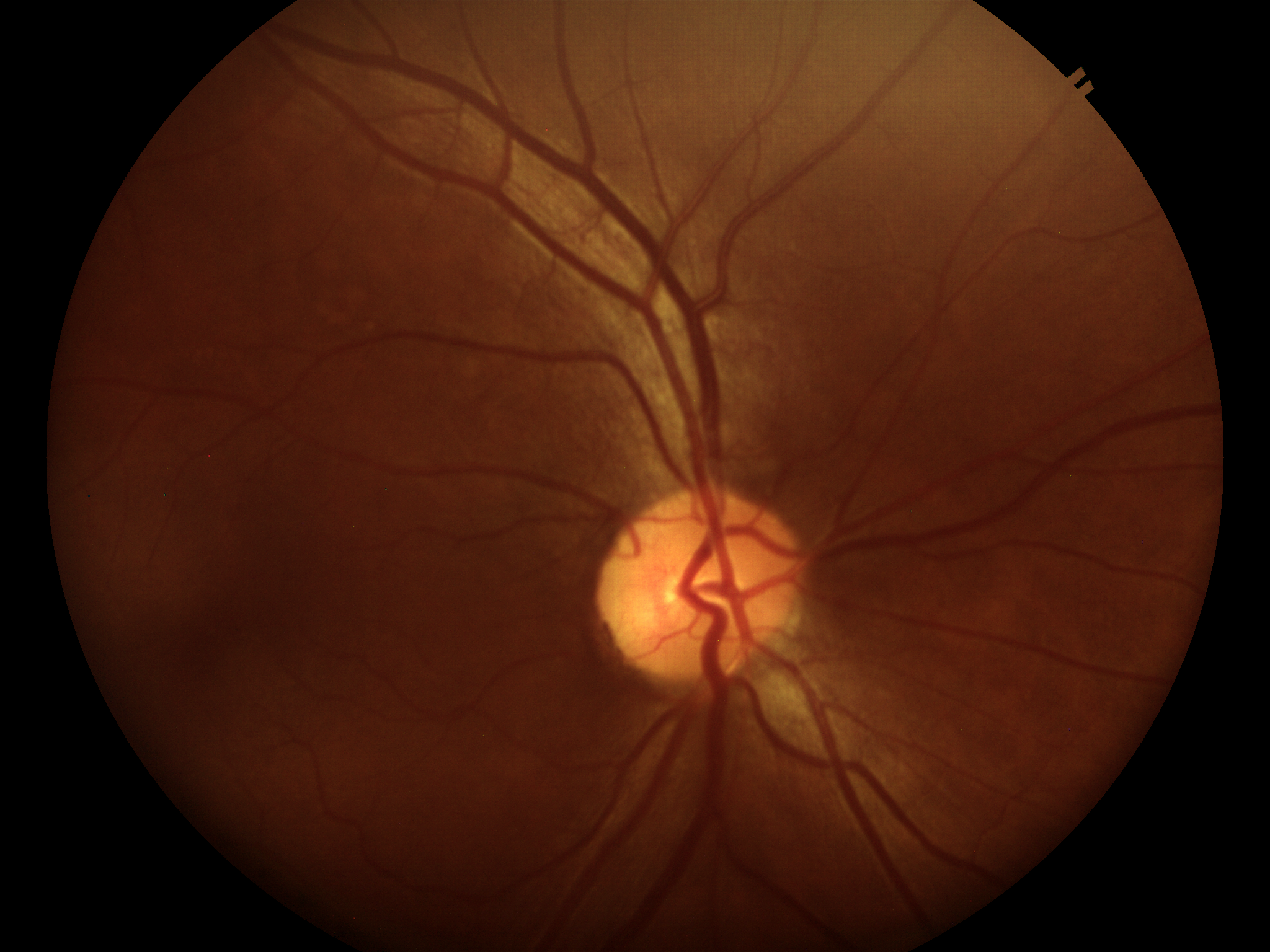}
      \caption{With Forus camera}     
    \end{subfigure}
    \begin{subfigure}{.3\columnwidth}
        \centering
         \includegraphics[width=\linewidth]{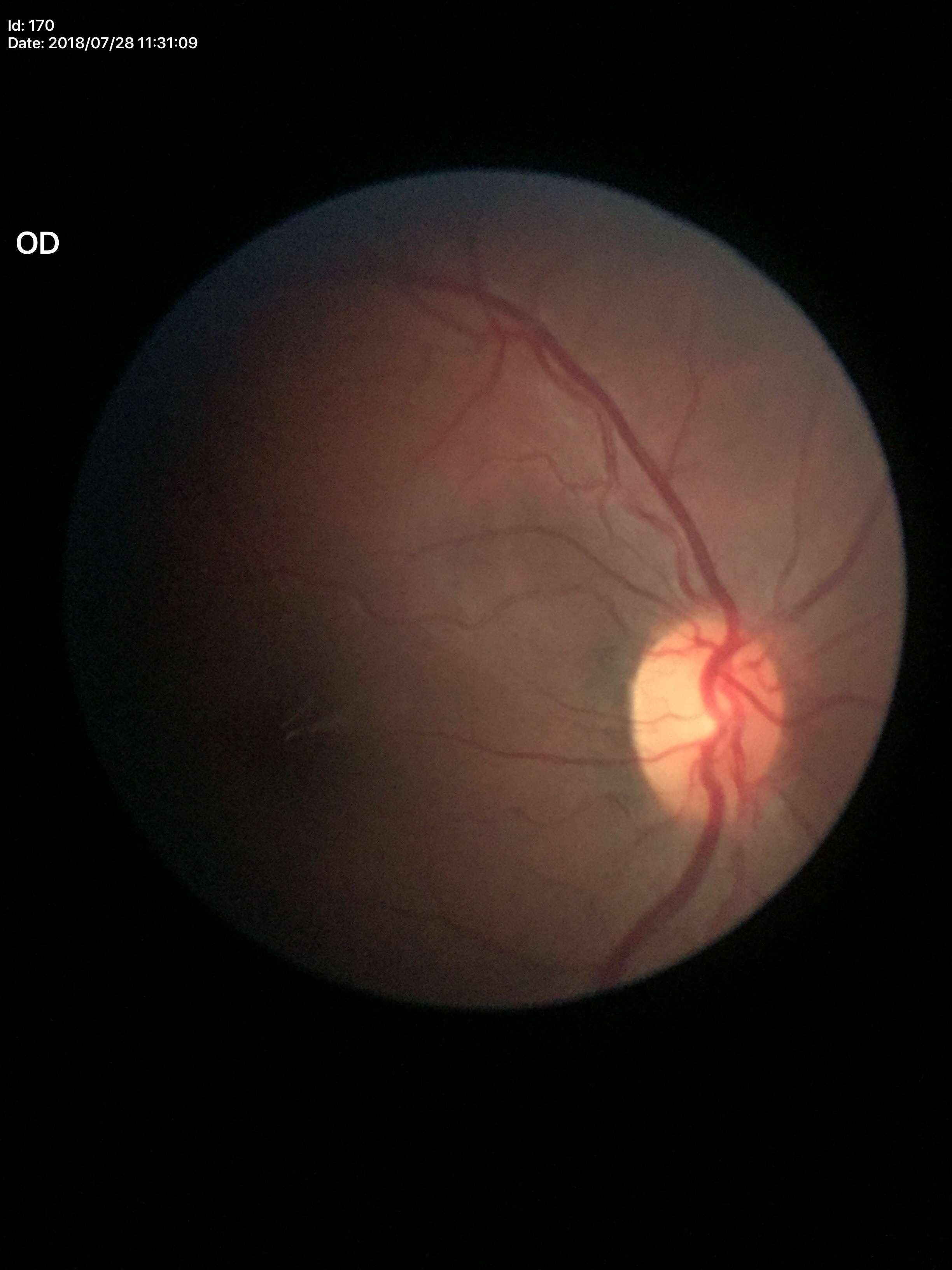}
      \caption{With Remidio camera}     
    \end{subfigure}
    \caption{Examples of eye fundus from each camera in the Chákṣu dataset}
    \label{fig:data_chaksu}
\end{figure}

Our second task is the binary classification of glaucoma in colour eye fundus images. We use the Chákṣu dataset~\cite{kumar2023chaksu}\footnote{Downloaded from \url{https://figshare.com/articles/dataset/Ch_k_u_A_glaucoma_specific_fundus_image_database/20123135}} which contains 1345 images acquired with three different fundus cameras as in Fig.~\ref{fig:data_chaksu}. Similar to the choice of PadChest, we chose Chákṣu as it was recently released and therefore less studied than other eye-fundus datasets. The dataset was labelled by five experts and contains, in addition to the glaucoma decision, the segmentation map of the optic disc and optic cup which are the structures used by clinicians to perform the diagnosis. Multiple fusion techniques are available to combine the annotations: using the mean, the median, the majority and the Simultaneous Truth and Performance Level Estimation (STAPLE) algorithm~\cite{warfield2004simultaneous}. Dataset providers showed that fusing using the STAPLE algorithm gives the best alignment with the experts, we therefore used the masks from this fusion in our study. We evaluate OOD performance with data from the AIROGS challenge~\cite{de2023airogs}\footnote{Downloaded from \url{https://doi.org/10.5281/zenodo.5793241}} and, as the test set of the challenge is not accessible, we decided to use the first 18,000 images of the training set.

\subsubsection{Pre-processing:}
The pre-processing is similar for all datasets. We resized the images to $512\times512$ and normalized them. If needed, we added black padding to an image to preserve the aspect ratio of the original content after the resize.

\subsection{Experimental setting, masking of ROI and evaluation}
For the PadChest dataset, we split the images into a train and test set with 80/20 proportions. The testing set was already created in the Chákṣu dataset.
For both PadChest and Chákṣu, we train a Densenet-121 model~\cite{huang2017densely} pre-trained on imagenet-1k with all weights frozen except the last dense block and the classification head using a 5-fold cross-validation protocol. The final set of hyperparameters for both datasets can be seen in Table~\ref{table:training_hyperparameters}, all models of the same dataset use the same set of hyperparameters regardless of the masking strategy.
Five versions of the dataset are used to train five models: 
\begin{enumerate}
    \item \textit{Full} images
    \item Keeping only the outside of mask boundaries referred to as "\textit{No lungs/disc}"
    \item Keeping only the outside of a bounding box of the mask referred to as "\textit{No lungs/disc BB}"
    \item Keeping only the inside of the precise mask boundaries referred to as "\textit{Only lungs/disc}"
    \item Keeping only the inside a bounding box of the mask referred to as "\textit{Only lungs/disc BB}"
\end{enumerate}

\begin{figure}[H]
    \centering
    \begin{subfigure}{\columnwidth}
        \centering
        \includegraphics[width=\columnwidth,keepaspectratio]{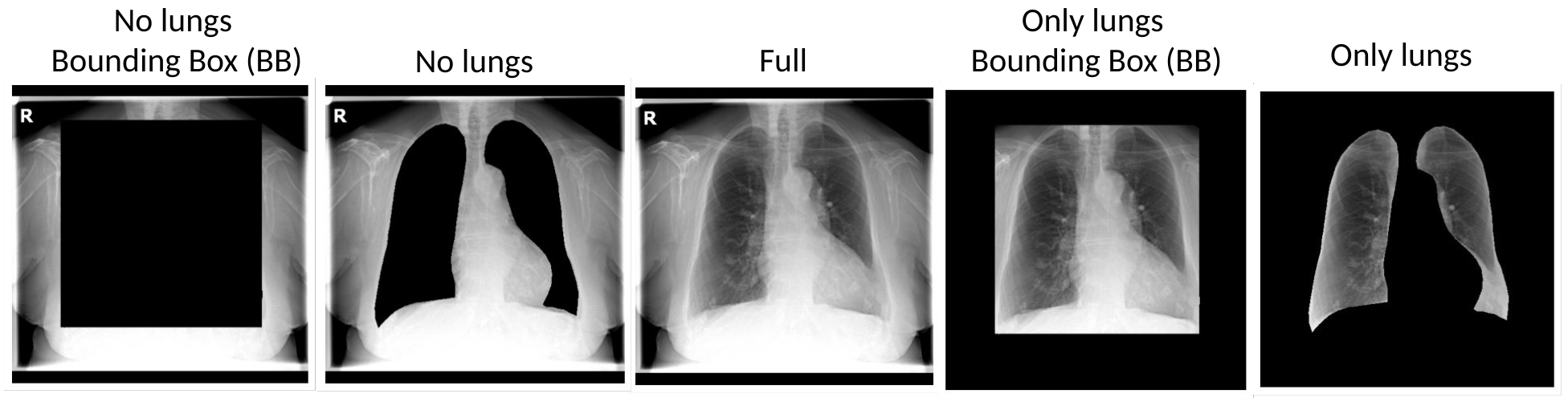}
        \caption{Chest X-ray from PadChest}
        \label{fig:cxr_data}
    \end{subfigure}
    \begin{subfigure}{\columnwidth}
        \centering
        \includegraphics[width=\columnwidth,keepaspectratio]{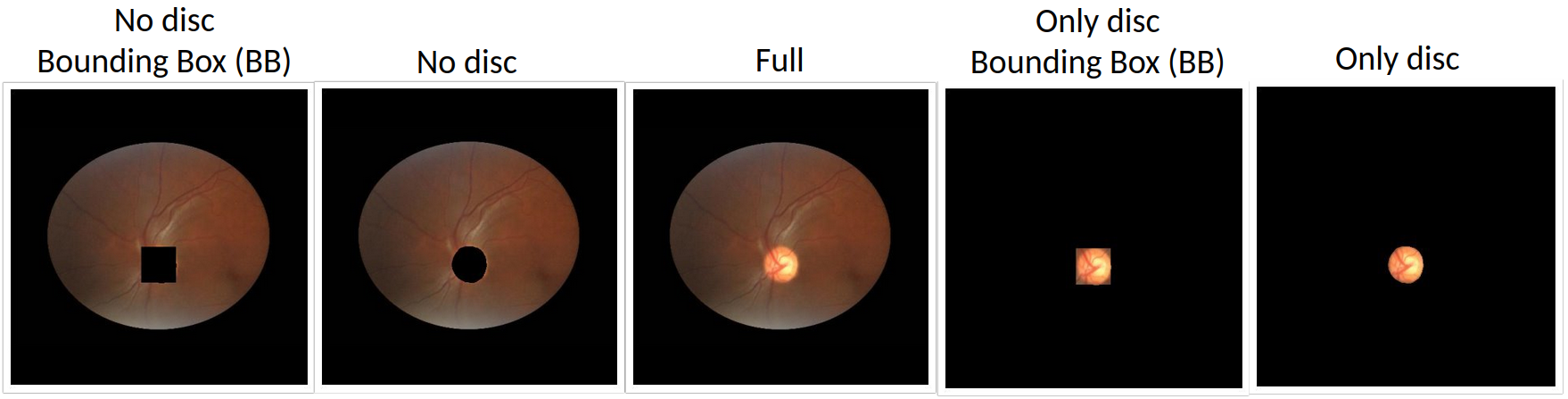}
        \caption{Eye fundus from Chákṣu dataset}
        \label{fig:eyefundus_data}
    \end{subfigure}
    \caption{Example of all masking strategies used in our study. Each strategy is used to train a separate model}
    \label{fig:masking_strategies}
\end{figure}

The usage of a bounding box is important to remove the shape information but also to include or exclude relevant parts like the heart for the cardiomegaly condition.
See Fig.~\ref{fig:masking_strategies} for an example of all strategies on a chest X-ray and an eye fundus image.
The models are evaluated using the Area Under the Curve (AUC) as this metric quantifies the ability of models to distinguish positive instances regardless of a chosen decision threshold.

\begin{table}[H]
\centering
\begin{tabular}{|p{0.32\textwidth}|p{0.32\textwidth}|p{0.32\textwidth}|}
\hline
Dataset        & PadChest             & Chákṣu               \\ \hline
Learning rate  & 1e-5                 & 1e-4                 \\ \hline
Batch size     & 32                   & 32                   \\ \hline
Loss           & Cross-entropy        & Weighted cross-entropy \\ \hline
Maximum epochs & 250                  & 250                  \\ \hline
Early stopping & 10 epochs, delta 1e-3 & 10 epochs, delta 1e-3 \\ \hline
Data augmentation & Random rotation (+/- 45°)\newline Random horizontal flip (p=0.5)\newline Brightness modification (factor 0.7,1.1) & Random rotation (+/- 45°)\newline Random horizontal flip (p=0.5)\newline Brightness modification (factor 0.7,1.1) \\ \hline

\end{tabular}
\caption{Final training hyperparameters for each dataset. The hyperparameters are the same for all masking strategies of a dataset}
\label{table:training_hyperparameters}
\end{table}

As a baseline, we train logistic regression classifiers using tabular data for the Padchest dataset using the same folds as for the image classification models but removing the samples for which at least one of the features was absent. 
For the chest x-rays, we train the models with the patient's birth year, the patient's sex and the projection view. 
No patient metadata was available for the eye fundus dataset to conduct such experiments. 
Baseline models on tabular data are recommended to evaluate the additive value of images compared to metadata~\cite{jimenez2025picture}. 

We evaluated the statistical significance of the differences between the AUCs of the different models with the Delong test~\cite{delong1988comparing,sun2014fast}. The Delong test is the most appropriate test to compare correlated AUC and was for example used in \cite{cheplygina2014classification,larsen2025new} to compare classifiers on medical images. As we have five models per version of the dataset, we concluded that a model was significantly better if the p-values were below 0.05 for at least three folds.

\subsection{Comparison of embeddings}
In addition to the results of the classification task, we analyse the generated embeddings to evaluate their similarity. 
We use the embedding vectors of dimension 1024 of the flattened global average pooling layer before the classification head of the model trained with \textit{full} images applied to images with different masking.
We use the t-SNE algorithm~\cite{van2008visualizing} to have a global view of the proximity of the different embeddings. The t-SNE algorithm transforms high-dimensional data into lower-dimension (often two or three dimensions to visualize the data) while preserving the local structure of the data in high-dimension.
We also compute the cosine similarity between the embeddings of full images and their different masked variation for a quantitative measure.

\subsection{Explainability method}
We use SHAP (SHapley Additive exPlanations)~\cite{lundberg2017shap} which divides the image into superpixels and computes the fair contribution of each superpixel using the change in the classifier outputs after their alteration using occlusion, blurring or inpainting method. We used an occlusion mask and set the number of evaluations to 1000.While other works pointed some limitations to its usage~\cite{klein2024navigating,bilodeau2024impossibility}, we decided to use SHAP following the results of Sun et al.~\cite{sun2023right}. They show its better performance on medical images containing shortcuts compared to gradient-based explanation methods such as Grad-Cam~\cite{selvaraju2017grad}.

\subsection{Study with a domain expert}

We compare the ability of the AI models and a domain expert to distinguish healthy and non-healthy images to assess whether it was possible to make the correct diagnosis for clinically relevant reasons in the images. It is for example useful for the effusion class which appears in the pleural space and could therefore be visible in images without lungs depending on the masks.
We asked a radiology resident with five years of experience and who has been involved previously in multiple AI projects to annotate images from each condition and the different masking strategies. 
The study was performed in two steps. 
During the first step, the radiology resident had to annotate ten images randomly selected from our training set and could ask any question. This phase was used to present the study to the radiology resident, to ask his opinion on the annotation tool and to improve it.

During the second step, we selected three images per condition and masking (75 images in total) in the testing set. We selected the images based on the output of the model: one image from the highest probabilities output, one near the median and one among the low probabilities. We informed the radiology resident that an image could contain one or multiple of the conditions but also conditions we do not track or not contain conditions at all. 
The radiology resident did not have access to other information such as the projection (Posterior-Anterior (PA) and Anterior-Posterior (AP)) which is important for some conditions such as cardiomegaly or pleural effusion as it impacts the visual aspect.

\section{Results}
\subsection{Different impact between chest X-rays and eye fundus}\label{sec:results_auc}

\begin{figure}[H]
    \centering
    \begin{subfigure}{.45\columnwidth}
        \centering
         \includegraphics[width=\linewidth]{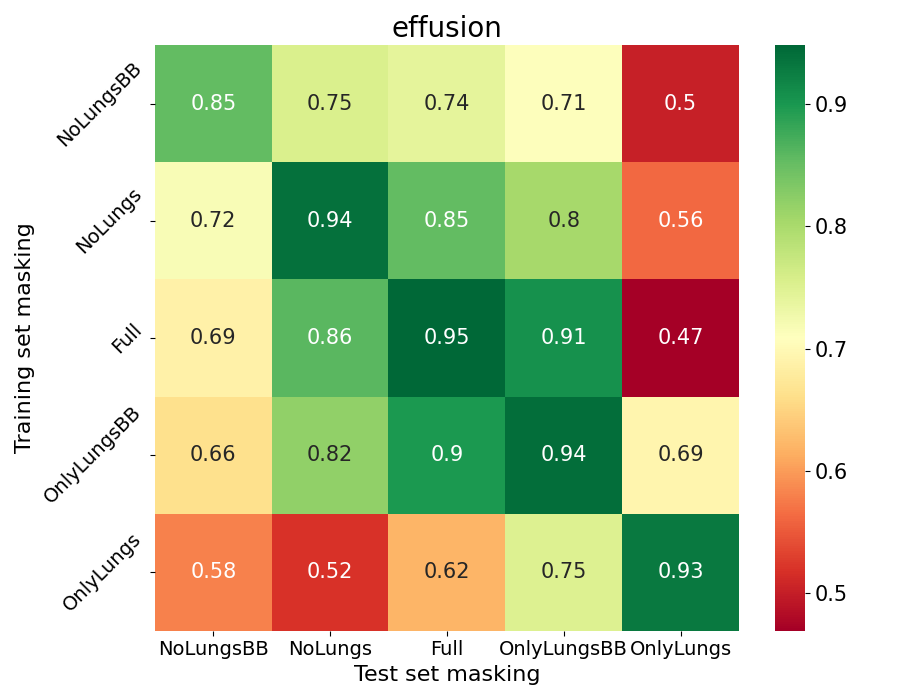}
      \caption{Mean AUC for the effusion class in PadChest}
      \label{subfig:mean_auc_padchest}
    \end{subfigure}
    \begin{subfigure}{.45\columnwidth}
      \centering
      \includegraphics[width=\linewidth]{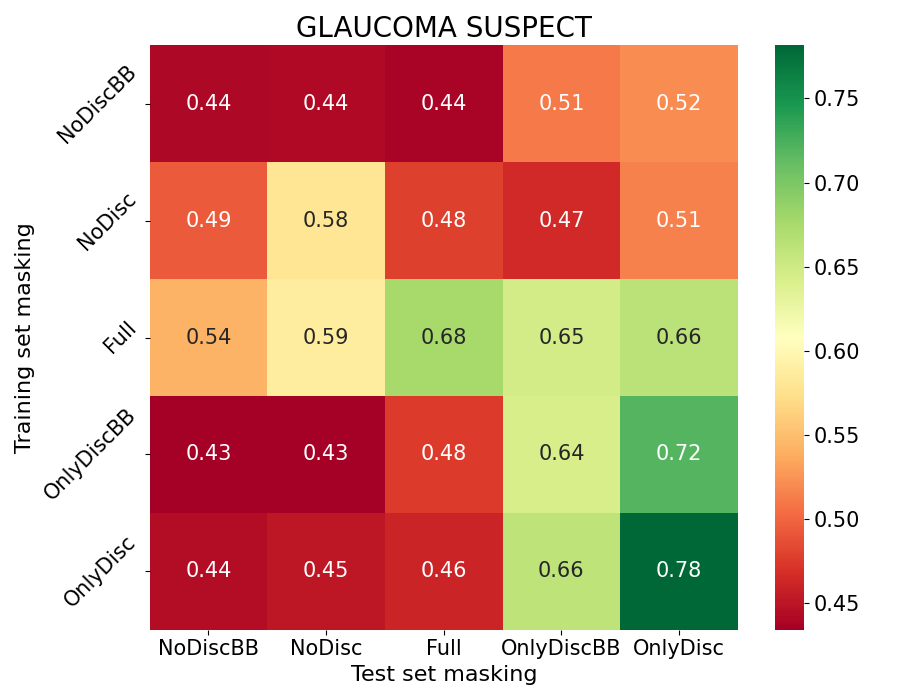}
        \caption{Mean AUC for the glaucoma class in Chákṣu}
        \label{subfig:mean_auc_chaksu}
    \end{subfigure}
    \caption{Mean AUC across the five models from 5-fold cross-validation on the testing set with different masking for training and evaluation images. The x-axis shows the masking strategy of the images used to evaluate the model, the y-axis shows the masking strategy of the images used to train the model. The color range in both figures is different to fit the range of the specific set.}
    \label{fig:mean_auc}
\end{figure}

We observed different results between chest X-rays and eye fundus. For example, Fig.~\ref{fig:mean_auc} shows the AUC for the effusion class of the PadChest dataset and glaucoma of the Chákṣu dataset.
Results for all classes of PadChest are consistent and can be found in Appendix~\ref{appendix:auc}. 

On the PadChest dataset, we found that every model is able to obtain good performances when trained and evaluated on similar masking even on images without lungs using the bounding box which removes a high proportion of the image (AUC between 0.85 and 0.93 on the effusion class and minimum 0.74 on the pneumonia with models trained on \textit{No lungs BB}). These values are all above the baseline results obtained with the logistic regression classifiers on tabular data (mean AUC: 0.71 atelectasis, 0.77 cardiomegaly, 0.72 effusion, 0.60 pneumonia, 0.70 pneumothorax), it is however interesting to note that these AUCs are also above the random value of 0.5. Another interesting result is that for all the classes, models trained on \textit{Full} images performed better when evaluated on images without lungs than on images with only lungs. These results show that the models are less able to distinguish images when keeping only the lungs with the precise masks with even some classes near or below the random value (atelectasis, effusion, and pneumonia). However, we see an important increase when using the bounding box which may indicate the usage of elements near the lungs captured when using the bounding box. It is worth noting that depending on the receptive field of the model, features at the boundaries of the lungs in \textit{Only lungs} images may be impacted by the masking.

\begin{figure}[H]
    \centering
    \begin{subfigure}{.45\columnwidth}
        \centering
         \includegraphics[width=\linewidth]{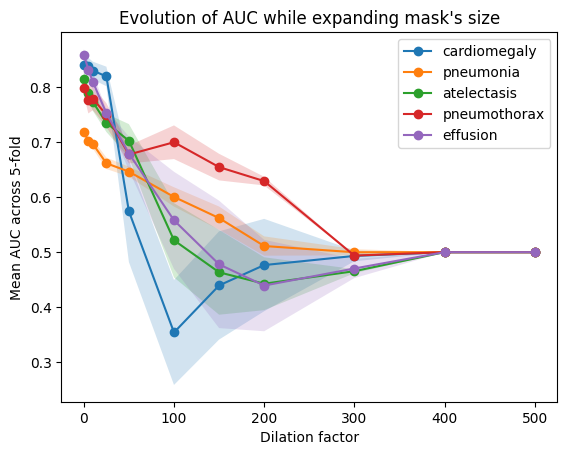}
      \caption{Images without lungs}
      \label{subfig:auc_dilation_NoLungs}
    \end{subfigure}
    \begin{subfigure}{.45\columnwidth}
      \centering
      \includegraphics[width=\linewidth]{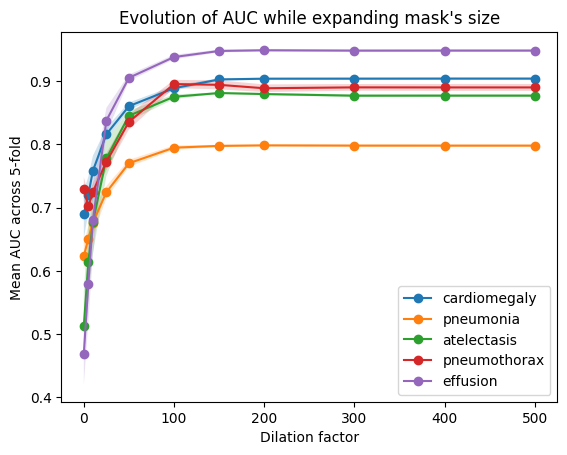}
        \caption{Images with only lungs}
        \label{subfig:auc_dilation_OnlyLungs}
    \end{subfigure}
    \caption{Evolution of the AUC with standard deviation of the models trained on \textit{Full} images applied to (a) images with only lungs and (b) images without the lungs while dilating the mask.}
    \label{fig:auc_dilation_lungs}
\end{figure}

Fig~\ref{fig:auc_dilation_lungs} shows the evolution of the AUC when increasing the mask size for images with only the lungs and without lungs applied to the model trained of \textit{Full} images. Fig.~\ref{fig:dilation_lungs} in Appendix~\ref{appendix:dilation} shows examples of images obtained at each dilation factor. For most classes, the highest drop of performance for images without lungs happens at a dilation factor of 100 or higher when most of the image is masked out. It is interesting to note that despite a drop, the AUC for the pneumonia class remains above random until completely masking out the images. 
For images with only the lungs we can see that a dilation factor of 50 is needed before achieving results close to \textit{Full} images. 
As we concluded from the AUCs using the bounding boxes, these results may reveal the usage of elements near the lungs but also sometimes at the border of the image.

The combination of these results, consistent over all classes, reinforced our hypothesis of models affected by spurious correlation in the dataset.

On the Chákṣu dataset, we can see random or near-random results on images without the region of interest with an 0.58 AUC for models trained and evaluated on \textit{No disc} and 0.44 for models trained and evaluated on \textit{No disc BB}. On the other hand, we obtain better performances for \textit{Full} images (0.68 AUC), \textit{Only disc BB} (0.64 AUC) and \textit{Only disc} (0.78 AUC). These performances are more aligned with what we would expect from models not affected by shortcut learning. 

To further test whether models could suffer from shortcut learning, we also assess how models are affected by the optic disc size. In practice, the cup-to-disc ratio or the rim-to-disc ratio are indicators of glaucoma but the size of the disc alone is not~\cite{coan2023automatic}. 

We evaluate the evolution of AUC when increasing the size of the masks for both \textit{Only disc} and \textit{No disc}. In this experiment, we either only increase the size of the masks for healthy images or glaucoma images. 
Fig.~\ref{fig:dilation_eye} in Appendix~\ref{appendix:dilation} shows example of both type of image at different dilation factor.

\begin{figure}[H]
    \centering
    \begin{subfigure}{.45\columnwidth}
        \centering
         \includegraphics[width=\linewidth]{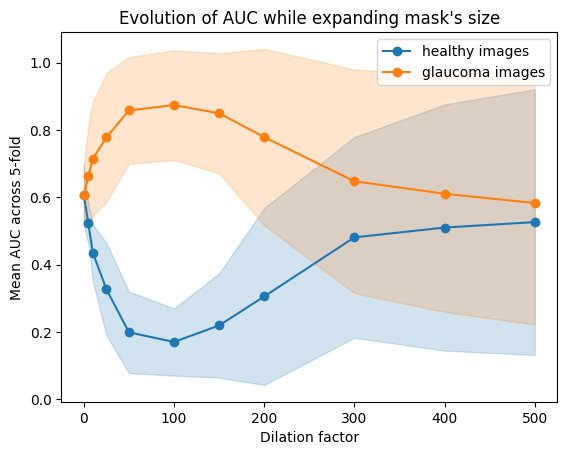}
      \caption{Trained on images without the optic disc (\textit{No disc})}     
    \end{subfigure}
    \begin{subfigure}{.45\columnwidth}
      \centering
      \includegraphics[width=\linewidth]{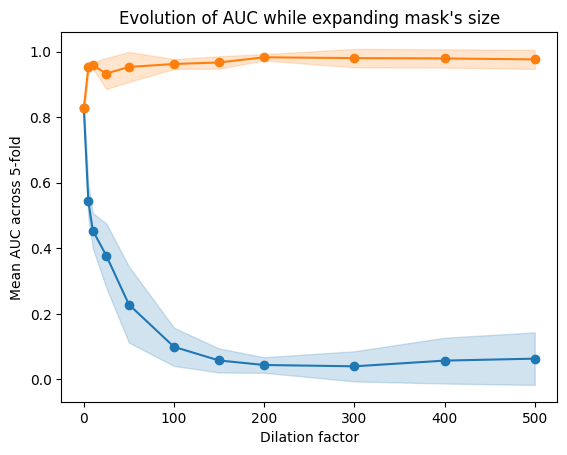}
        \caption{Trained on images without the outside of the optic disc (\textit{Only disc})}
    \end{subfigure}
    \caption{Evolution of the AUC with standard deviation when increasing mask size of healthy images (in blue) and glaucoma images (in orange)}
    \label{fig:evolution_auc_dilation}
\end{figure}

The results in Fig.~\ref{fig:evolution_auc_dilation} show that disk size impacts the models' outputs. For the \textit{No disc} model, the AUC increases for larger masks of glaucoma images until a dilation factor of 150 when it starts decreasing. For the model trained on \textit{Only disc}, the AUC reaches 1.0 at the smallest dilation factor of 5 and never decreases even at the highest value of 500.   
Note that this seems to also be a problem among clinicians as smaller discs may be more misclassified as healthy and larger discs as glaucoma~\cite{heijl1993optic}. Moreover, as the optic disc size varies across populations~\cite{od1999optic}, such spurious correlation could therefore lead to degraded performance in underrepresented populations.

\subsection{Poor out-of-distribution performance}

We hypothesised that the models trained without the region of interest would have lower performances on external datasets as shortcuts are often dataset-specific, relying on such shortcuts would therefore not transfer well to other datasets. 
Results in Table~\ref{table:ood} seem to partly align with our intuition. We indeed find poor performances of models trained without the ROI, almost always close to an AUC of 0.5. 
However, we can see that the results of models trained with only the ROI are often also close to this value. For cardiomegaly, effusion, pneumonia and glaucoma, the best performances are achieved by models trained on \textit{Full} images. Moreover, we observe no statistically significant difference between \textit{Full} and \textit{Only lung} on atelectasis for 3/5 folds. However, the results remain well below the results on the training dataset. 
This result confirms the poor transferability across populations and the need to test and fine-tune models on the target population before applying the models in a different context like a different hospital.

\begin{table}[H]
\centering
\begin{tabular}{l|lllll|l}
\hline
Masking     & Atelectasis           & Cardiomegaly          & Effusion              & Pneumonia             & Pneumothorax          & Glaucoma \\ \hline
No ROI BB   & 0.53\tiny{+/-0.004}          & 0.54\tiny{+/-0.026}          & 0.58\tiny{+/-0.005}          & 0.52\tiny{+/-0.009}          & 0.54\tiny{+/-0.016}          &0.49\tiny{+/-0.045}          \\
No ROI     & 0.52\tiny{+/-0.005}          & 0.54\tiny{+/-0.022}          & 0.63\tiny{+/-0.004}          & 0.55\tiny{+/-0.018}          & 0.51\tiny{+/-0.020}          &0.52\tiny{+/-0.050}          \\ \hline
Full      & 0.59\tiny{+/-0.004}          & \textbf{0.72\tiny{+/-0.016}*} & \textbf{0.72\tiny{+/-0.002}*} & \textbf{0.62\tiny{+/-0.012}*} & 0.56\tiny{+/-0.016}          &\textbf{0.65\tiny{+/-0.048}*}          \\ \hline
Only ROI BB & \textbf{0.60\tiny{+/-0.002}} & 0.67\tiny{+/-0.012}          & 0.67\tiny{+/-0.009}          & 0.59\tiny{+/-0.011}          & 0.51\tiny{+/-0.011}          &0.52\tiny{+/-0.030}          \\
Only ROI   & 0.54\tiny{+/-0.013}          & 0.58\tiny{+/-0.022}          & 0.53\tiny{+/-0.021}          & 0.55\tiny{+/-0.014}          & \textbf{0.60\tiny{+/-0.029}*} &0.53\tiny{+/-0.031}          \\ \hline
\end{tabular}
\caption{Out-of-distribution evaluation results - Mean AUC (+/- standard deviation) of 5-fold models per masking type and class on full images of the ChestX-ray14 dataset's test set (Atelectasis, Cardiomegaly, Effusion, Pneumonia and Pneumothorax) and AIROGS dataset (Glaucoma). * indicates statistical significance for three or more folds (p-values < 0.05)}
\label{table:ood}
\end{table}

\subsection{Closer embeddings between full images and without ROI}

\begin{figure}[H]
    \centering
    \begin{subfigure}{.45\columnwidth}
        \centering
        \includegraphics[width=\linewidth]{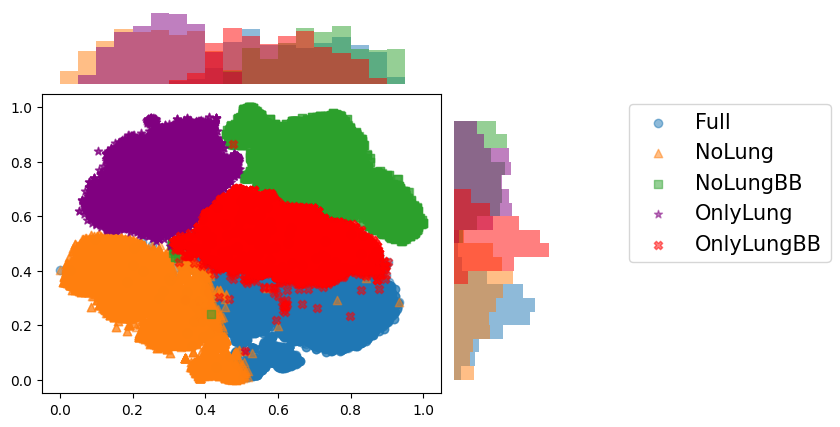}
        \caption{PadChest}     
        \label{fig:tsne_padchest}
    \end{subfigure}
    \begin{subfigure}{.45\columnwidth}
      \centering
      \includegraphics[width=\linewidth]{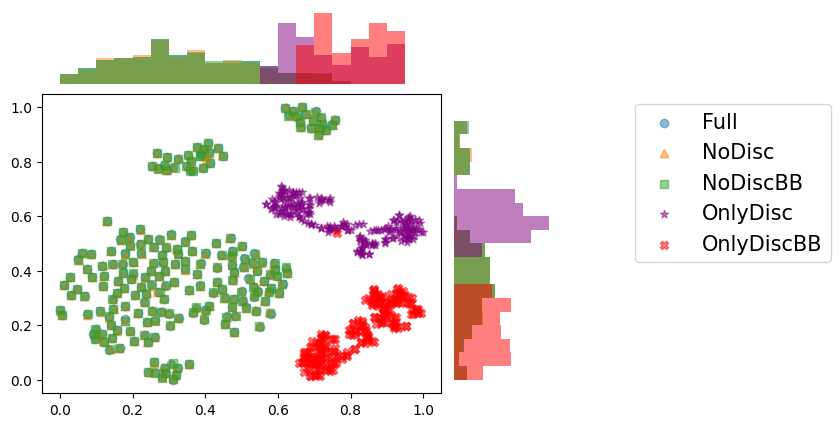}
        \caption{Chákṣu}
        \label{fig:tsne_chaksu}
    \end{subfigure}
    \caption{t-SNE of the embeddings before classification head of the model trained with the full images dataset on images with various masking from (a) PadChest dataset and (b) the Chákṣu dataset. In blue circles are full images, in orange triangles without ROI using the mask, in green squares without the ROI using the bounding box, in purple stars with only the ROI using the bounding box and in red crosses only the ROI using the mask}
    \label{fig:tsne}
\end{figure}

We show the results from the t-SNE algorithm on the PadChest dataset in Fig.~\ref{fig:tsne_padchest}. We find that the embeddings of each masking type clearly cluster together with a small overlapping between embeddings of the full images, images without lungs and images with only lungs using the bounding box. 
On the Chákṣu dataset in Fig.~\ref{fig:tsne_chaksu}, we see the embeddings of \textit{Full images}, \textit{No disc} and \textit{No disc BB} completely overlapping while \textit{Only disc} and \textit{Only disc BB} have a clear cluster.

For chest X-rays, the mean cosine similarity in Table~\ref{table:embedding_similarity} is higher when comparing full images with images without lungs using masks, followed by images with only the lungs using the bounding box, then images without lungs using the bounding box and finally, the lowest similarity is with images without lungs using masks. Both similarities of \textit{No lungs} and \textit{Only lungs BB} are close, which could also confirm the usage of elements at the border or between the lungs that are not included in the images of \textit{Only lungs} using the mask.

For eye fundus, the results in Table~\ref{table:embedding_similarity_eyefundus} are aligned with the t-SNE and show the high proximity of images without the ROI and \textit{Full} images compared to images with only the ROI. 
It however doesn't align with the AUC measure in Fig.~\ref{subfig:mean_auc_chaksu} as we would expect AUC of \textit{Full} images to be closer to No ROI images than Only ROI images. 
The close embeddings before the classification head may be caused by the similarity of base images as \textit{Full} and \textit{No lungs} images share more common pixels than \textit{Full} and \textit{Only lungs}. 

These results show that despite being commonly used, the interpretation of t-SNE visualisations and cosine similarities of features before the classification head is limited, should be considered carefully, and completed with further analysis.

\begin{table}[H]
\centering
\begin{tabular}{lcccccc}
Image label       & All & Cardiomegaly & Pneumonia & Atelectasis & Pneumothorax & Effusion \\
\midrule
No lungs     & 0.90{\tiny+/-0.02}            & 0.90{\tiny+/-0.02}            & 0.90{\tiny+/-0.02}          & 0.90{\tiny+/-0.02}            & 0.91{\tiny+/-0.02}             & 0.90{\tiny+/-0.02}          \\
No lungs BB   & 0.83{\tiny+/-0.04}             & 0.83{\tiny+/-0.04}             & 0.85{\tiny+/-0.04}          & 0.86{\tiny+/-0.04}             & 0.85{\tiny+/-0.04}           & 0.85{\tiny+/-0.04}         \\
Only lungs   & 0.76{\tiny+/-0.03}             & 0.74{\tiny+/-0.03}             & 0.76{\tiny+/-0.04}          &  0.75{\tiny+/-0.03}           & 0.79{\tiny+/-0.02}             & 0.73{\tiny+/-0.03}        \\
Only lungs BB & 0.89{\tiny+/-0.03}             & 0.88{\tiny+/-0.03}             & 0.88{\tiny+/-0.04}          &  0.87{\tiny+/-0.04}           &  0.89{\tiny+/-0.04}            & 0.86{\tiny+/-0.04}          \\
\bottomrule
\end{tabular}
\caption{Mean cosine similarity (+/- standard deviation) over pairs of images of the first fold, between embeddings of images with different masking by the model trained on full images of the PadChest dataset}
\label{table:embedding_similarity}
\end{table}

\begin{table}[H]
\centering
\begin{tabular}{lcc}
Image label       & All & Glaucoma                                      \\
\midrule
No disc     & 0.97{\tiny+/- 0.01}           & 0.97{\tiny+/- 0.01}       \\
No disc BB   & 0.98{\tiny+/- 0.01}            & 0.98{\tiny+/- 0.01}     \\
Only disc   & 0.69{\tiny+/- 0.03}         & 0.71{\tiny+/- 0.04}         \\
Only disc BB & 0.68{\tiny+/- 0.03}           & 0.69{\tiny+/- 0.03}      \\
\bottomrule
\end{tabular}
\caption{Mean cosine similarity (+/- standard deviation) over images of the first fold, between embeddings of images with different masking by the model trained on full images of the Chákṣu dataset}
\label{table:embedding_similarity_eyefundus}
\end{table}

\subsection{Models can use both the relevant structure and the shortcut}

\begin{figure}[H]
    \centering
    \begin{subfigure}{.45\columnwidth}
        \centering
        \includegraphics[width=\columnwidth,keepaspectratio]{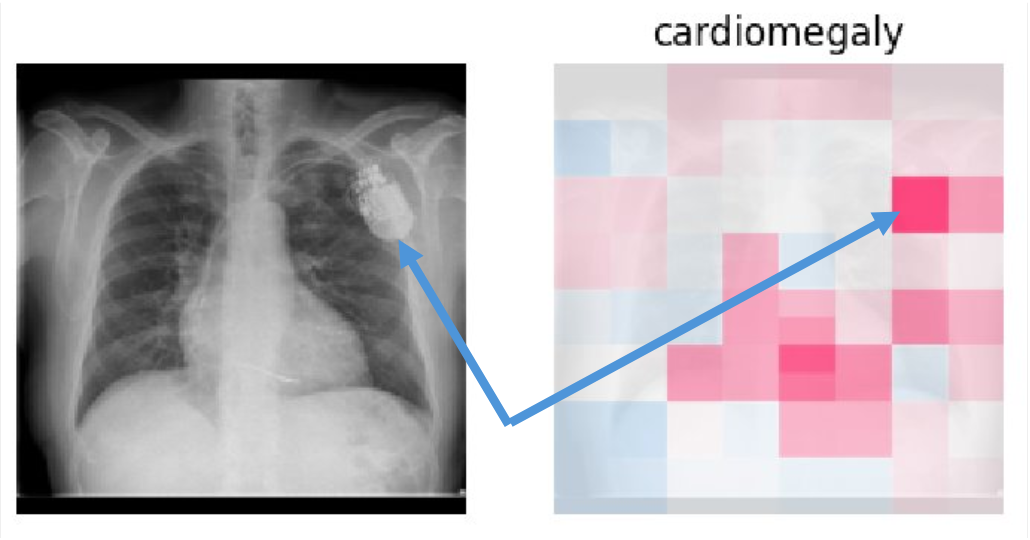}
        \caption{With pacemaker}
        \label{fig:shap_results_pacemaker}
    \end{subfigure}
    \begin{subfigure}{.45\columnwidth}
        \centering
        \includegraphics[width=\columnwidth,keepaspectratio]{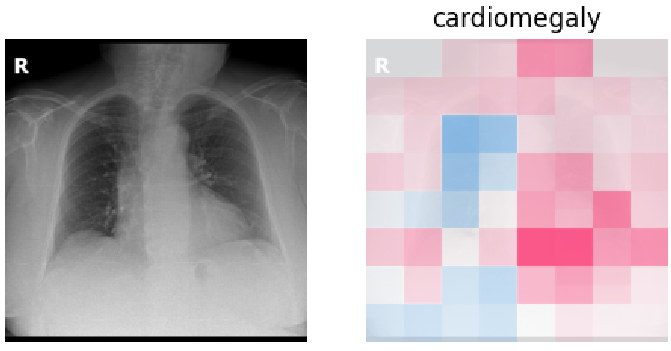}
        \caption{Without pacemaker}
        \label{fig:shap_results_nopacemaker}
    \end{subfigure}
    \begin{subfigure}{.45\columnwidth}
        \centering
        \includegraphics[width=\columnwidth,keepaspectratio]{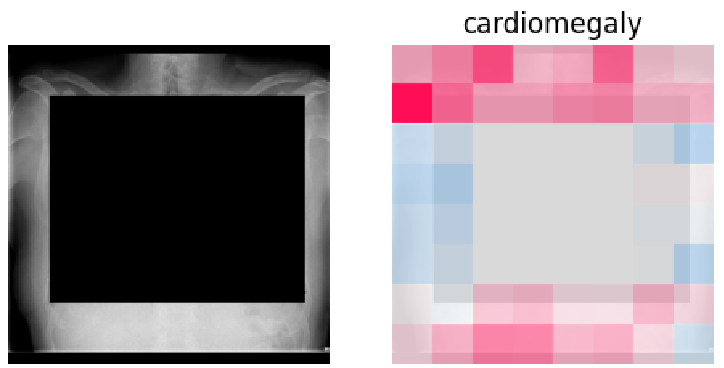}
        \caption{Without the lungs}
        \label{fig:shap_results_cardiomegaly_noBB}
    \end{subfigure}
    
    \caption{SHAP values of the cardiomegaly class with and without a pacemaker, and without the lungs region. Red values indicate parts increasing the probability. The blue arrows in (a) show the pacemaker location.}
    \label{fig:shap_results_cardiomegaly}
    
\end{figure}

Fig.~\ref{fig:shap_results_cardiomegaly} shows examples of behaviours found with the explainability maps from SHAP on the PadChest dataset. 
We can for example see the usage of pacemakers in the decision-making for the cardiomegaly class (Fig.~\ref{fig:shap_results_pacemaker}) even though the model still uses the relevant part of the image (i.e. the heart) in the decision-making (Fig.~\ref{fig:shap_results_nopacemaker}). It is interesting to note that while there can be a correlation between cardiomegaly and pacemakers there is no causal link between the two.  
Similarly to tubes used as shortcuts for pneumothorax classification~\cite{jimenez2023detecting,oakden2020hidden}, this shows an example of shortcuts that can be easily detected and are medically understandable. 
On the other hand, explainability maps of images without the ROI highlight the usage of elements at the edge of the image even far from the region of interest (e.g. the heart for cardiomegaly such as in Fig.~\ref{fig:shap_results_cardiomegaly_noBB}). In this case, the regions highlighted by SHAP don't present any apparent reason for their usage. It is therefore difficult to provide a clear explanation of the model output. Moreover, SHAP provides a local explanation of the output and may therefore not be able to detect possible global shortcuts such as specific noise created by different scanners or settings during the acquisition.
These results highlight an important distinction between the non-obvious shortcuts, such as noise, and the medically understandable shortcuts, such as pacemakers, that can be more easily detected and therefore mitigated. 

\begin{figure}[H]
    \centering
    \includegraphics[width=0.60\columnwidth,keepaspectratio]{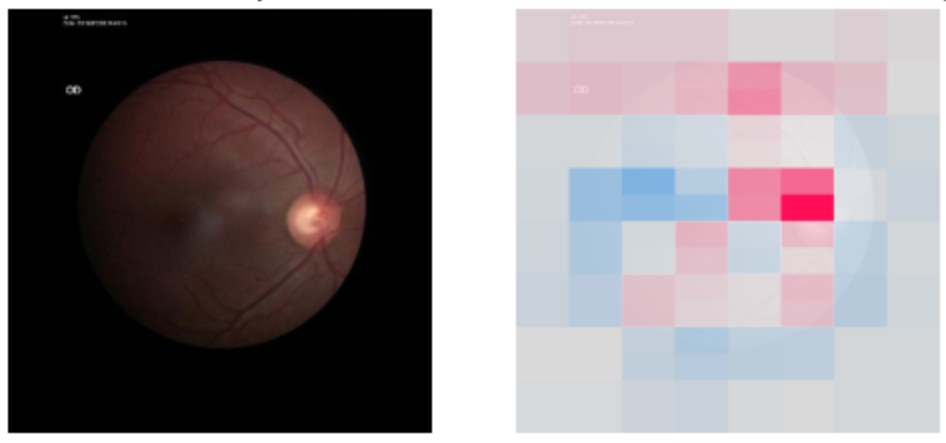}
    \caption{SHAP values of an image with glaucoma}
    \label{fig:shap_results_chaksu}
\end{figure}

For the Chákṣu dataset, we can also see in Fig.~\ref{fig:shap_results_chaksu} the usage of eye boundaries in the decision, these areas contain differences for the 3 cameras with for example the text in the Remedio images or the cropped boundaries in the Forus images. SHAP also highlights the optic disc but the finding in Section~\ref{sec:results_auc} put doubt on the correct usage of this area by the model as it could also be only looking at the disc size and not the cup-to-disc ratio. These two results highlight the importance of not only relying on explainability methods to evaluate models' behaviour but also conducting more thorough analyses. 

\subsection{Lack of ROI and context makes the diagnosis difficult for a domain expert}

\begin{figure}[H]
    \centering
    \includegraphics[width=0.5\columnwidth,keepaspectratio]{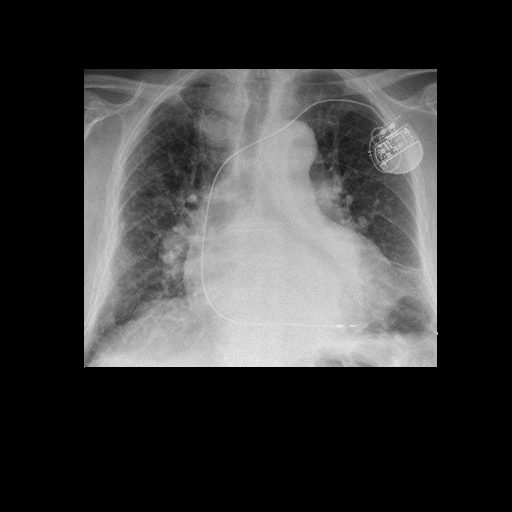}
    \caption{One of the images labelled by the radiologist with the comment "The diagnosis would be heavily dependent on whether it is AP or PA. Also potential effusion would be hard to exclude on AP"}
    \label{fig:annotation_APPA}
\end{figure}

Out of the 35 cases of the conditions present in images without the ROI (using the mask or bounding box) only 2 were found by the radiology resident, both in images with lungs removed using the mask. One was a patient with cardiomegaly which is diagnosed using the heart still present when only removing the lungs. 
This result strengthens our hypothesis that the model relies on non-relevant features to domain experts.
Results on full images and images containing only the ROI also reveal the difficulty of the task when only relying on the image. A recurring observation across images is the lack of information about the projection as in Fig.~\ref{fig:annotation_APPA}. 

\begin{figure}[H]
    \centering
    \begin{subfigure}{.45\columnwidth}
        \centering
        \includegraphics[width=\columnwidth,keepaspectratio]{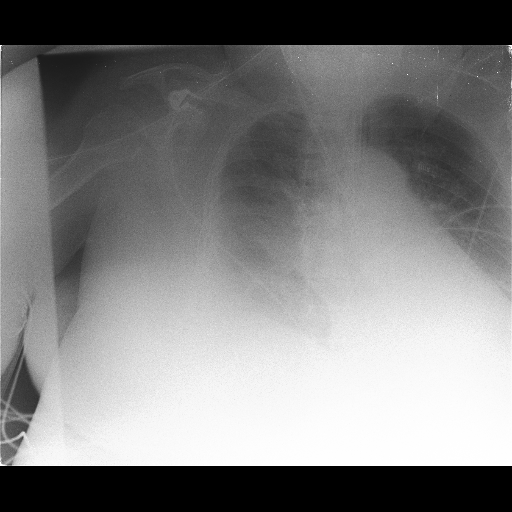}
        \caption{Image with poor quality.}
        \label{fig:bad_quality}
    \end{subfigure}
    \begin{subfigure}{.45\columnwidth}
        \centering
        \includegraphics[width=\columnwidth,keepaspectratio]{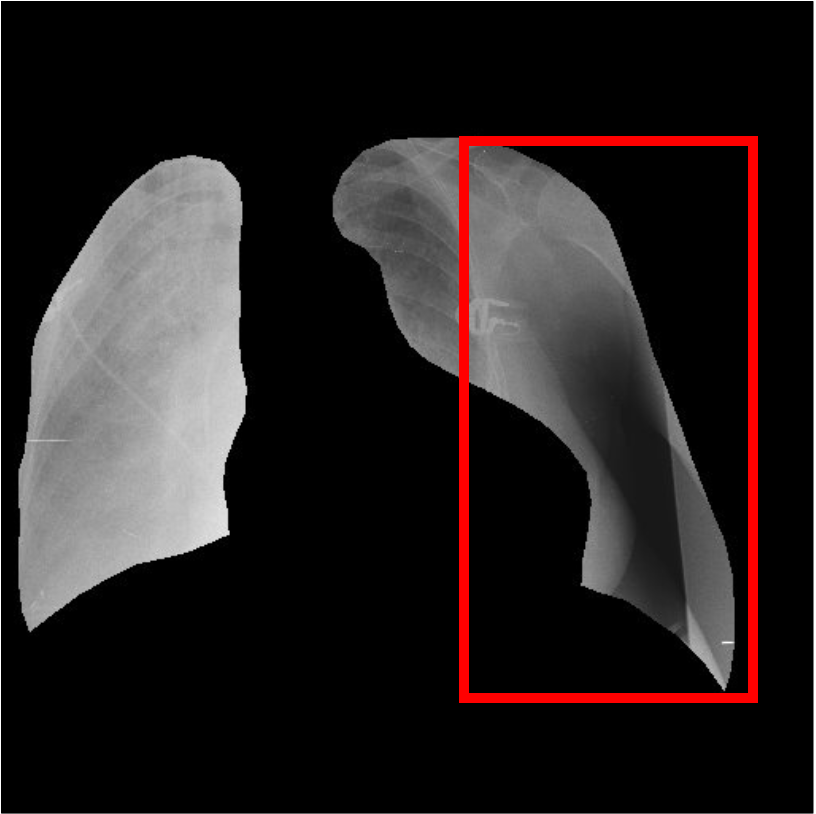}
        \caption{Image with a non-accurate mask.}
        \label{fig:bad_mask}
    \end{subfigure}
    \caption{Example of images with either a bad quality making the diagnosis difficult for a radiologist due to the blur in the whole image or with a non-accurate mask indicated by the red box outside the chest area.}
    \label{fig:bad_images}
\end{figure}

Due to the automatic selection of images based on probabilities, we also found cases in Fig.~\ref{fig:bad_images} of quality problems both in the original and the mask dataset. 
This highlights one of the problems of using large-scale public datasets for which the assessment of quality is difficult but essential as shown on other tasks and datasets to obtain better and more robust models~\cite{aubreville2024cleaning,abhishek2025investigating}.

\section{Discussion}
We studied how convolutional neural networks are affected by the masking of different parts of medical images to highlight the effect of shortcut learning.
We analyzed performances, comparing the AUC of each masking strategy, but also went beyond by using an explainability method and comparing the embeddings.
Our experiments showed the ability of CNN models to rely on non-relevant parts of medical images. 
Our results highlight the difficulty of detecting such behaviours. The AUC obtained on different masking configurations may show strong evidence of shortcut learning, like in chest X-rays. However, it can also be more hidden like in glaucoma detection for which the performance analysis reveals no signs of shortcut learning. 
With eye-fundus photography, we also showcased a limitation of relying on the explainability method as the model could be focusing on a correct region but not using the correct feature. It also shows the potential difficulty of detecting patient-level shortcuts described in \cite{juodelyte2024source} such as anatomical confounders when they are located at the area of interest.

The results from the study with the radiology resident highlight how relying on a single source of information (e.g. an image) makes the diagnosis more difficult or even impossible in some cases. In this setting, the lack of context may guide the model to rely more on shortcuts (such as pacemakers). The recent development of multimodal models~\cite{schouten2024navigating} may therefore help mitigate these issues as the model has access to more clinically relevant information, although it is yet unknown whether the additional modalities could introduce other shortcuts. 
Importantly, for certain pathologies, chest X-rays are only one part of the diagnostic process, for example a chest X-ray could be used for triage before a follow-up CT scan. 

We only used a single CNN architecture, a Densenet-121, for our experiments. Other works showed the effect of shortcut learning on other CNN architectures in medical imaging~\cite{boland2024there,hill2024risk}. 
Other architectures such as transformers or vision-language models are becoming popular in medical image analysis, but their robustness to shortcuts has only been studied in the natural images domain~\cite{ghosal2024vision}. 
As our code repository allows us to easily adapt the framework to other models, it would be of interest to apply it to these different types of architectures in future works.

We focused on the discriminative capacity of models using the AUC as the metric. However, other aspects, such as calibration or fairness across different subgroups, need to be taken into account during the evaluation, especially if the aim is to deploy the model in a clinical environment \cite{maierhein2024metrics}. 
Finally, our method only provides a way to evaluate the risk of shortcut learning, further analysis may be performed to conclude on its absence or presence and other techniques for improving data quality and/or mitigating the model bias would therefore need to be applied.

We believe that the analysis of potential biases should be an important step in the evaluation of AI models. In addition to the poor transferability to other domains such as another hospital, these spurious correlations can have a major impact on the diagnosis of underrepresented populations. 
We also believe that our field would benefit from going beyond widely used performance metrics such as the AUC that don't necessarily translate into real clinical progress. 
To this aim, collaboration with hospitals and domain experts is essential to ground research in more clinical knowledge, ensuring better data understanding but also the relevancy and added value of developed algorithms from a clinical point of view.

Both our and other recent works show the need to question whether models the community trains actually learn salient characteristics of the diseases. It is likely that published models overfit to spurious features, and should not be applied in practice until the community figures out how to resolve these issues and achieve improved robustness.

\subsubsection{Acknowledgments} 
\ifdefined\DOUBLEBLIND
***
\else
We thank Casper Anton Poulsen, Trine Naja Eriksen and Cathrine Damgaard for their early work within this research line. We also thank Evangelia Christodoulou for her advice on the statistical significance test.
This work was supported by the DFF - Inge Lehmann 1134-00017B and Novo Nordisk Foundation NNF21OC0068816.
\fi

%
%

\bibliographystyle{unsrt}
\bibliography{refs}

\begin{thebibliography}{10}

\bibitem{Wu2023ClinicalAdoption}
Kevin Wu, Eric Wu, Brandon Theodorou, Weixin Liang, Christina Mack, Lucas Glass, Jimeng Sun, and James Zou.
\newblock Characterizing the clinical adoption of medical ai devices through u.s. insurance claims.
\newblock {\em NEJM AI}, 1(1):AIoa2300030, 2023.

\bibitem{shen2019artificial}
Jiayi Shen, Casper~JP Zhang, Bangsheng Jiang, Jiebin Chen, Jian Song, Zherui Liu, Zonglin He, Sum~Yi Wong, Po-Han Fang, Wai-Kit Ming, et~al.
\newblock Artificial intelligence versus clinicians in disease diagnosis: systematic review.
\newblock {\em JMIR medical informatics}, 7(3):e10010, 2019.

\bibitem{plesner2023autonomous}
Louis~L Plesner, Felix~C M{\"u}ller, Janus~D Nybing, Lene~C Laustrup, Finn Rasmussen, Olav~W Nielsen, Mikael Boesen, and Michael~B Andersen.
\newblock Autonomous chest radiograph reporting using ai: estimation of clinical impact.
\newblock {\em Radiology}, 307(3):e222268, 2023.

\bibitem{Wu2021DeviceEvaluation}
Eric Wu, Kevin Wu, Roxana Daneshjou, David Ouyang, Daniel~E. Ho, and James Zou.
\newblock How medical ai devices are evaluated: limitations and recommendations from an analysis of fda approvals.
\newblock {\em Nature Medicine}, 27(4):582–584, 2021.

\bibitem{jimenez2023detecting}
Amelia Jim{\'e}nez-S{\'a}nchez, Dovile Juodelyte, Bethany Chamberlain, and Veronika Cheplygina.
\newblock Detecting shortcuts in medical images-a case study in chest x-rays.
\newblock In {\em 2023 IEEE 20th International Symposium on Biomedical Imaging (ISBI)}, pages 1--5, 2023.

\bibitem{oakden2020hidden}
Lauren Oakden-Rayner, Jared Dunnmon, Gustavo Carneiro, and Christopher R{\'e}.
\newblock Hidden stratification causes clinically meaningful failures in machine learning for medical imaging.
\newblock In {\em ACM Conference on Health, Inference, and Learning}, pages 151--159, 2020.

\bibitem{roschewitz2024counterfactual}
M{\'e}lanie Roschewitz, Fabio de~Sousa~Ribeiro, Tian Xia, Galvin Khara, and Ben Glocker.
\newblock Counterfactual contrastive learning: robust representations via causal image synthesis.
\newblock In {\em MICCAI Workshop on Data Engineering in Medical Imaging}, pages 22--32, 2024.

\bibitem{boland2024all}
Christopher Boland, Owen Anderson, Keith~A Goatman, John Hipwell, Sotirios~A Tsaftaris, and Sonia Dahdouh.
\newblock All you need is a guiding hand: Mitigating shortcut bias in deep learning models for medical imaging.
\newblock In {\em MICCAI Workshop on Fairness of AI in Medical Imaging}, pages 67--77, 2024.

\bibitem{liz2023deep}
Helena Liz, Javier Huertas-Tato, Manuel S{\'a}nchez-Monta{\~n}{\'e}s, Javier Del~Ser, and David Camacho.
\newblock Deep learning for understanding multilabel imbalanced chest x-ray datasets.
\newblock {\em Future Generation Computer Systems}, 144:291--306, 2023.

\bibitem{de2023airogs}
Coen De~Vente, Koenraad~A Vermeer, Nicolas Jaccard, He~Wang, Hongyi Sun, Firas Khader, Daniel Truhn, Temirgali Aimyshev, Yerkebulan Zhanibekuly, Tien-Dung Le, et~al.
\newblock Airogs: artificial intelligence for robust glaucoma screening challenge.
\newblock {\em IEEE transactions on medical imaging}, 2023.

\bibitem{kassem2021machine}
Mohamed~A Kassem, Khalid~M Hosny, Robertas Dama{\v{s}}evi{\v{c}}ius, and Mohamed~Meselhy Eltoukhy.
\newblock Machine learning and deep learning methods for skin lesion classification and diagnosis: a systematic review.
\newblock {\em Diagnostics}, 11(8):1390, 2021.

\bibitem{sabour2017dynamic}
Sara Sabour, Nicholas Frosst, and Geoffrey~E Hinton.
\newblock Dynamic routing between capsules.
\newblock {\em Advances in neural information processing systems}, 30, 2017.

\bibitem{jimenez2018capsule}
Amelia Jim{\'e}nez-S{\'a}nchez, Shadi Albarqouni, and Diana Mateus.
\newblock Capsule networks against medical imaging data challenges.
\newblock In {\em Intravascular Imaging and Computer Assisted Stenting and Large-Scale Annotation of Biomedical Data and Expert Label Synthesis}, pages 150--160, 2018.

\bibitem{dos2020capsule}
Patrick Ryan~Sales Dos~Santos, Vit{\'o}ria de~Carvalho~Brito, Antonio~Oseas de~Carvalho~Filho, Fl{\'a}vio Henrique~Duarte de~Ara{\'u}jo, Ricardo de Andrade~Lira Rab{\^e}lo, and Mano~Joseph Mathew.
\newblock A capsule network-based for identification of glaucoma in retinal images.
\newblock In {\em 2020 IEEE Symposium on Computers and Communications (ISCC)}, pages 1--6, 2020.

\bibitem{shamshad2023transformers}
Fahad Shamshad, Salman Khan, Syed~Waqas Zamir, Muhammad~Haris Khan, Munawar Hayat, Fahad~Shahbaz Khan, and Huazhu Fu.
\newblock Transformers in medical imaging: A survey.
\newblock {\em Medical Image Analysis}, 88:102802, 2023.

\bibitem{marikkar2023lt}
Umar Marikkar, Sara Atito, Muhammad Awais, and Adam Mahdi.
\newblock Lt-vit: A vision transformer for multi-label chest x-ray classification.
\newblock In {\em 2023 IEEE International Conference on Image Processing (ICIP)}, pages 2565--2569, 2023.

\bibitem{fan2023detecting}
Rui Fan, Kamran Alipour, Christopher Bowd, Mark Christopher, Nicole Brye, James~A Proudfoot, Michael~H Goldbaum, Akram Belghith, Christopher~A Girkin, Massimo~A Fazio, et~al.
\newblock Detecting glaucoma from fundus photographs using deep learning without convolutions: transformer for improved generalization.
\newblock {\em Ophthalmology science}, 3(1):100233, 2023.

\bibitem{moutakanni2024advancing}
Th{\'e}o Moutakanni, Piotr Bojanowski, Guillaume Chassagnon, C{\'e}line Hudelot, Armand Joulin, Yann LeCun, Matthew Muckley, Maxime Oquab, Marie-Pierre Revel, and Maria Vakalopoulou.
\newblock Advancing human-centric ai for robust x-ray analysis through holistic self-supervised learning.
\newblock {\em arXiv preprint arXiv:2405.01469}, 2024.

\bibitem{wu2023towards}
Chaoyi Wu, Xiaoman Zhang, Ya~Zhang, Yanfeng Wang, and Weidi Xie.
\newblock Towards generalist foundation model for radiology.
\newblock {\em arXiv preprint arXiv:2308.02463}, 2023.

\bibitem{liu2024visual}
Chunyu Liu, Yixiao Jin, Zhouyu Guan, Tingyao Li, Yiming Qin, Bo~Qian, Zehua Jiang, Yilan Wu, Xiangning Wang, Ying~Feng Zheng, et~al.
\newblock Visual--language foundation models in medicine.
\newblock {\em The Visual Computer}, pages 1--20, 2024.

\bibitem{li2024visionunite}
Zihan Li, Diping Song, Zefeng Yang, Deming Wang, Fei Li, Xiulan Zhang, Paul~E Kinahan, and Yu~Qiao.
\newblock Visionunite: A vision-language foundation model for ophthalmology enhanced with clinical knowledge.
\newblock {\em arXiv preprint arXiv:2408.02865}, 2024.

\bibitem{geirhos2020shortcut}
Robert Geirhos, J{\"o}rn-Henrik Jacobsen, Claudio Michaelis, Richard Zemel, Wieland Brendel, Matthias Bethge, and Felix~A Wichmann.
\newblock Shortcut learning in deep neural networks.
\newblock {\em Nature Machine Intelligence}, 2(11):665--673, 2020.

\bibitem{banerjee2023shortcuts}
Imon Banerjee, Kamanasish Bhattacharjee, John~L Burns, Hari Trivedi, Saptarshi Purkayastha, Laleh Seyyed-Kalantari, Bhavik~N Patel, Rakesh Shiradkar, and Judy Gichoya.
\newblock “shortcuts” causing bias in radiology artificial intelligence: causes, evaluation and mitigation.
\newblock {\em Journal of the American College of Radiology}, 2023.

\bibitem{vasquez2024detecting}
Constanza V{\'a}squez-Venegas, Chenwei Wu, Saketh Sundar, Renata Pr{\^o}a, Francis~Joshua Beloy, Jillian~Reeze Medina, Megan McNichol, Krishnaveni Parvataneni, Nicholas Kurtzman, Felipe Mirshawka, et~al.
\newblock Detecting and mitigating the clever hans effect in medical imaging: A scoping review.
\newblock {\em Journal of Imaging Informatics in Medicine}, pages 1--17, 2024.

\bibitem{wargnier2024unexpected}
Valentine Wargnier-Dauchelle, Thomas Grenier, and Micha{\"e}l Sdika.
\newblock An unexpected confounder: how brain shape can be used to classify mri scans?
\newblock In {\em Medical Imaging with Deep Learning}, 2024.

\bibitem{boland2024there}
Christopher Boland, Keith~A Goatman, Sotirios~A Tsaftaris, and Sonia Dahdouh.
\newblock There are no shortcuts to anywhere worth going: Identifying shortcuts in deep learning models for medical image analysis.
\newblock In {\em Medical Imaging with Deep Learning}, 2024.

\bibitem{weng2025fast}
Nina Weng, Paraskevas Pegios, Eike Petersen, Aasa Feragen, and Siavash Bigdeli.
\newblock Fast diffusion-based counterfactuals for shortcut removal and generation.
\newblock In {\em European Conference on Computer Vision}, pages 338--357, 2025.

\bibitem{perez2024radedit}
Fernando P{\'e}rez-Garc{\'\i}a, Sam Bond-Taylor, Pedro~P Sanchez, Boris van Breugel, Daniel~C Castro, Harshita Sharma, Valentina Salvatelli, Maria~TA Wetscherek, Hannah Richardson, Matthew~P Lungren, et~al.
\newblock Radedit: stress-testing biomedical vision models via diffusion image editing.
\newblock In {\em European Conference on Computer Vision}, pages 358--376, 2024.

\bibitem{bissoto2019constructing}
Alceu Bissoto, Michel Fornaciali, Eduardo Valle, and Sandra Avila.
\newblock (de)constructing bias on skin lesion datasets.
\newblock In {\em Proceedings of the IEEE/CVF Conference on Computer Vision and Pattern Recognition (CVPR) Workshops}, 2019.

\bibitem{hemelings2021deep}
Ruben Hemelings, Bart Elen, Jo{\~a}o Barbosa-Breda, Matthew~B Blaschko, Patrick De~Boever, and Ingeborg Stalmans.
\newblock Deep learning on fundus images detects glaucoma beyond the optic disc.
\newblock {\em Scientific Reports}, 11(1):20313, 2021.

\bibitem{haynes2024generalisation}
Sophie~Crawford Haynes, Pamela Johnston, and Eyad Elyan.
\newblock Generalisation challenges in deep learning models for medical imagery: insights from external validation of covid-19 classifiers.
\newblock {\em Multimedia Tools and Applications}, 83(31):76753--76772, 2024.

\bibitem{aslani2022optimising}
Shahab Aslani, Watjana Lilaonitkul, Vaishnavi Gnanananthan, Divya Raj, Bojidar Rangelov, Alexandra~L Young, Yipeng Hu, Paul Taylor, Daniel~C Alexander, NCCID Collaborative, et~al.
\newblock Optimising chest x-rays for image analysis by identifying and removing confounding factors.
\newblock In {\em International Conference on Medical Imaging and Computer-Aided Diagnosis}, pages 245--254. Springer, 2022.

\bibitem{lin2024shortcut}
Manxi Lin, Nina Weng, Kamil Mikolaj, Zahra Bashir, Morten~BS Svendsen, Martin~G Tolsgaard, Anders~N Christensen, and Aasa Feragen.
\newblock Shortcut learning in medical image segmentation.
\newblock In {\em International Conference on Medical Image Computing and Computer-Assisted Intervention}, pages 623--633, 2024.

\bibitem{juodelyte2024source}
Dovile Juodelyte, Yucheng Lu, Amelia Jim{\'e}nez-S{\'a}nchez, Sabrina Bottazzi, Enzo Ferrante, and Veronika Cheplygina.
\newblock Source matters: Source dataset impact on model robustness in medical imaging.
\newblock In {\em International Workshop on Applications of Medical AI}, pages 105--115, 2024.

\bibitem{deng2009imagenet}
Jia Deng, Wei Dong, Richard Socher, Li-Jia Li, Kai Li, and Li~Fei-Fei.
\newblock Imagenet: A large-scale hierarchical image database.
\newblock In {\em Computer Vision and Pattern Recognition, 2009. CVPR 2009. IEEE Conference on}, pages 248--255, 2009.

\bibitem{mei2022radimagenet}
Xueyan Mei, Zelong Liu, Philip~M Robson, Brett Marinelli, Mingqian Huang, Amish Doshi, Adam Jacobi, Chendi Cao, Katherine~E Link, Thomas Yang, et~al.
\newblock Radimagenet: an open radiologic deep learning research dataset for effective transfer learning.
\newblock {\em Radiology: Artificial Intelligence}, 4(5):e210315, 2022.

\bibitem{sun2023right}
Susu Sun, Lisa~M Koch, and Christian~F Baumgartner.
\newblock Right for the wrong reason: Can interpretable ml techniques detect spurious correlations?
\newblock In {\em International Conference on Medical Image Computing and Computer-Assisted Intervention}, pages 425--434. Springer, 2023.

\bibitem{kim2024transparent}
Chanwoo Kim, Soham~U Gadgil, Alex~J DeGrave, Jesutofunmi~A Omiye, Zhuo~Ran Cai, Roxana Daneshjou, and Su-In Lee.
\newblock Transparent medical image ai via an image--text foundation model grounded in medical literature.
\newblock {\em Nature Medicine}, pages 1--12, 2024.

\bibitem{koh2020concept}
Pang~Wei Koh, Thao Nguyen, Yew~Siang Tang, Stephen Mussmann, Emma Pierson, Been Kim, and Percy Liang.
\newblock Concept bottleneck models.
\newblock In {\em International conference on machine learning}, pages 5338--5348, 2020.

\bibitem{Bustos2020PadChest}
Aurelia Bustos, Antonio Pertusa, Jose-Maria Salinas, and Maria de~la Iglesia-Vay{\'a}.
\newblock Padchest: A large chest x-ray image dataset with multi-label annotated reports.
\newblock {\em Medical Image Analysis}, 66:101797, 2020.

\bibitem{Irvin2019Chexpert}
Jeremy Irvin, Pranav Rajpurkar, Michael Ko, Yifan Yu, Silviana Ciurea-Ilcus, Chris Chute, Henrik Marklund, Behzad Haghgoo, Robyn Ball, Katie Shpanskaya, et~al.
\newblock Chexpert: A large chest radiograph dataset with uncertainty labels and expert comparison.
\newblock In {\em AAAI Conference on Artificial Intelligence}, volume~33, pages 590--597, 2019.

\bibitem{Wang2017chestxray8}
Xiaosong Wang, Yifan Peng, Le~Lu, Zhiyong Lu, Mohammadhadi Bagheri, and Ronald~M Summers.
\newblock Chestx-ray8: Hospital-scale chest x-ray database and benchmarks on weakly-supervised classification and localization of common thorax diseases.
\newblock In {\em Computer Vision and Pattern Recognition}, pages 2097--2106, 2017.

\bibitem{gaggion2024chexmask}
Nicol{\'a}s Gaggion, Candelaria Mosquera, Lucas Mansilla, Julia~Mariel Saidman, Martina Aineseder, Diego~H Milone, and Enzo Ferrante.
\newblock Chexmask: a large-scale dataset of anatomical segmentation masks for multi-center chest x-ray images.
\newblock {\em Scientific Data}, 11(1):511, 2024.

\bibitem{valindria2017reverse}
Vanya~V Valindria, Ioannis Lavdas, Wenjia Bai, Konstantinos Kamnitsas, Eric~O Aboagye, Andrea~G Rockall, Daniel Rueckert, and Ben Glocker.
\newblock Reverse classification accuracy: predicting segmentation performance in the absence of ground truth.
\newblock {\em IEEE transactions on medical imaging}, 36(8):1597--1606, 2017.

\bibitem{kumar2023chaksu}
JR~Harish Kumar, Chandra~Sekhar Seelamantula, JH~Gagan, Yogish~S Kamath, Neetha~IR Kuzhuppilly, U~Vivekanand, Preeti Gupta, and Shilpa Patil.
\newblock Ch{\'a}kṣu: A glaucoma specific fundus image database.
\newblock {\em Scientific data}, 10(1):70, 2023.

\bibitem{warfield2004simultaneous}
Simon~K Warfield, Kelly~H Zou, and William~M Wells.
\newblock Simultaneous truth and performance level estimation {(STAPLE)}: an algorithm for the validation of image segmentation.
\newblock {\em IEEE Transactions on Medical Imaging}, 23(7):903--921, 2004.

\bibitem{huang2017densely}
Gao Huang, Zhuang Liu, Laurens Van Der~Maaten, and Kilian~Q Weinberger.
\newblock Densely connected convolutional networks.
\newblock In {\em Proceedings of the IEEE conference on computer vision and pattern recognition}, pages 4700--4708, 2017.

\bibitem{jimenez2025picture}
Amelia Jim{\'e}nez-S{\'a}nchez, Natalia-Rozalia Avlona, Sarah de~Boer, V{\'\i}ctor~M Campello, Aasa Feragen, Enzo Ferrante, Melanie Ganz, Judy~Wawira Gichoya, Camila Gonz{\'a}lez, Steff Groefsema, et~al.
\newblock In the picture: Medical imaging datasets, artifacts, and their living review.
\newblock {\em arXiv preprint arXiv:2501.10727}, 2025.

\bibitem{delong1988comparing}
Elizabeth~R DeLong, David~M DeLong, and Daniel~L Clarke-Pearson.
\newblock Comparing the areas under two or more correlated receiver operating characteristic curves: a nonparametric approach.
\newblock {\em Biometrics}, pages 837--845, 1988.

\bibitem{sun2014fast}
Xu~Sun and Weichao Xu.
\newblock Fast implementation of delong’s algorithm for comparing the areas under correlated receiver operating characteristic curves.
\newblock {\em IEEE Signal Processing Letters}, 21(11):1389--1393, 2014.

\bibitem{cheplygina2014classification}
Veronika Cheplygina, Lauge S{\o}rensen, David~MJ Tax, Jesper~Holst Pedersen, Marco Loog, and Marleen De~Bruijne.
\newblock Classification of copd with multiple instance learning.
\newblock In {\em 2014 22nd International Conference on pattern recognition}, pages 1508--1513, 2014.

\bibitem{larsen2025new}
Kristoffer Larsen, Zhuo He, Fernando de~A~Fernandes, Xinwei Zhang, Chen Zhao, Qiuying Sha, Claudio~T Mesquita, Diana Paez, Ernest~V Garcia, Jiangang Zou, et~al.
\newblock A new method using deep learning to predict the response to cardiac resynchronization therapy.
\newblock {\em Journal of Imaging Informatics in Medicine}, pages 1--17, 2025.

\bibitem{van2008visualizing}
Laurens Van~der Maaten and Geoffrey Hinton.
\newblock Visualizing data using t-sne.
\newblock {\em Journal of machine learning research}, 9(11), 2008.

\bibitem{lundberg2017shap}
Scott~M Lundberg and Su-In Lee.
\newblock A unified approach to interpreting model predictions.
\newblock {\em Advances in neural information processing systems}, 30, 2017.

\bibitem{klein2024navigating}
Lukas Klein, Carsten L\"{u}th, Udo Schlegel, Till Bungert, Mennatallah El-Assady, and Paul Jaeger.
\newblock Navigating the maze of explainable ai: A systematic approach to evaluating methods and metrics.
\newblock In {\em Advances in Neural Information Processing Systems}, volume~37, pages 67106--67146, 2024.

\bibitem{bilodeau2024impossibility}
Blair Bilodeau, Natasha Jaques, Pang~Wei Koh, and Been Kim.
\newblock Impossibility theorems for feature attribution.
\newblock {\em Proceedings of the National Academy of Sciences}, 121(2):e2304406120, 2024.

\bibitem{selvaraju2017grad}
Ramprasaath~R Selvaraju, Michael Cogswell, Abhishek Das, Ramakrishna Vedantam, Devi Parikh, and Dhruv Batra.
\newblock Grad-cam: Visual explanations from deep networks via gradient-based localization.
\newblock In {\em Proceedings of the IEEE international conference on computer vision}, pages 618--626, 2017.

\bibitem{coan2023automatic}
Lauren~J Coan, Bryan~M Williams, Venkatesh~Krishna Adithya, Swati Upadhyaya, Ala Alkafri, Silvester Czanner, Rengaraj Venkatesh, Colin~E Willoughby, Srinivasan Kavitha, and Gabriela Czanner.
\newblock Automatic detection of glaucoma via fundus imaging and artificial intelligence: A review.
\newblock {\em Survey of ophthalmology}, 68(1):17--41, 2023.

\bibitem{heijl1993optic}
Anders Heijl and Harras M{\"o}lder.
\newblock Optic disc diameter influences the ability to detect glaucomatous disc damage.
\newblock {\em Acta ophthalmologica}, 71(1):122--129, 1993.

\bibitem{od1999optic}
Michael D~Hancox OD.
\newblock Optic disc size, an important consideration in the glaucoma evaluation.
\newblock {\em Clinical Eye and Vision Care}, 11(2):59--62, 1999.

\bibitem{aubreville2024cleaning}
Marc Aubreville, Jonathan Ganz, Jonas Ammeling, Christopher Kaltenecker, and Christof Bertram.
\newblock Model-based cleaning of the quilt-1m pathology dataset for text-conditional image synthesis.
\newblock In {\em Medical Imaging with Deep Learning}.

\bibitem{abhishek2025investigating}
Kumar Abhishek, Aditi Jain, and Ghassan Hamarneh.
\newblock Investigating the quality of dermamnist and fitzpatrick17k dermatological image datasets.
\newblock {\em Scientific Data}, 12(1):196, 2025.

\bibitem{schouten2024navigating}
Daan Schouten, Giulia Nicoletti, Bas Dille, Catherine Chia, Pierpaolo Vendittelli, Megan Schuurmans, Geert Litjens, and Nadieh Khalili.
\newblock Navigating the landscape of multimodal ai in medicine: a scoping review on technical challenges and clinical applications.
\newblock {\em arXiv preprint arXiv:2411.03782}, 2024.

\bibitem{hill2024risk}
Brandon~G Hill, Frances~L Koback, and Peter~L Schilling.
\newblock The risk of shortcutting in deep learning algorithms for medical imaging research.
\newblock {\em Scientific Reports}, 14(1):29224, 2024.

\bibitem{ghosal2024vision}
Soumya~Suvra Ghosal and Yixuan Li.
\newblock Are vision transformers robust to spurious correlations?
\newblock {\em International Journal of Computer Vision}, 132(3):689--709, 2024.

\bibitem{maierhein2024metrics}
Lena Maier-Hein, Annika Reinke, Patrick Godau, Minu~D Tizabi, Florian Buettner, Evangelia Christodoulou, Ben Glocker, Fabian Isensee, Jens Kleesiek, Michal Kozubek, et~al.
\newblock Metrics reloaded: recommendations for image analysis validation.
\newblock {\em Nature methods}, pages 1--18, 2024.

\end{thebibliography}

\newpage
\appendix
\section{AUC per label and masking}\label{appendix:auc}

\begin{figure}[H]
    \centering
    \begin{subfigure}{.45\columnwidth}
        \centering
         \includegraphics[width=\linewidth]{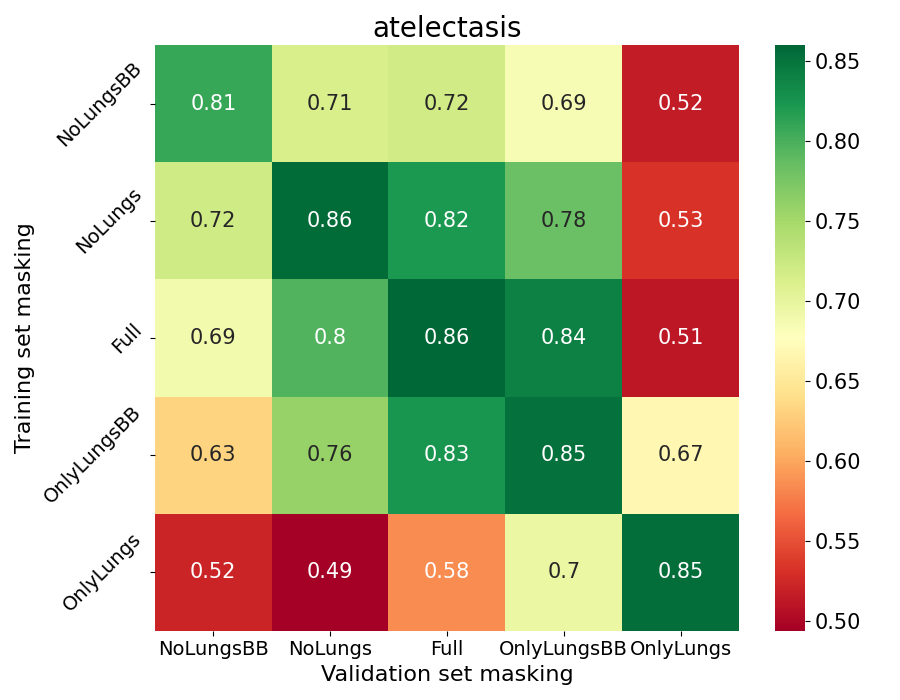}
      \caption{on 5-folds validation sets}     
    \end{subfigure}
    \begin{subfigure}{.45\columnwidth}
      \centering
      \includegraphics[width=\linewidth]{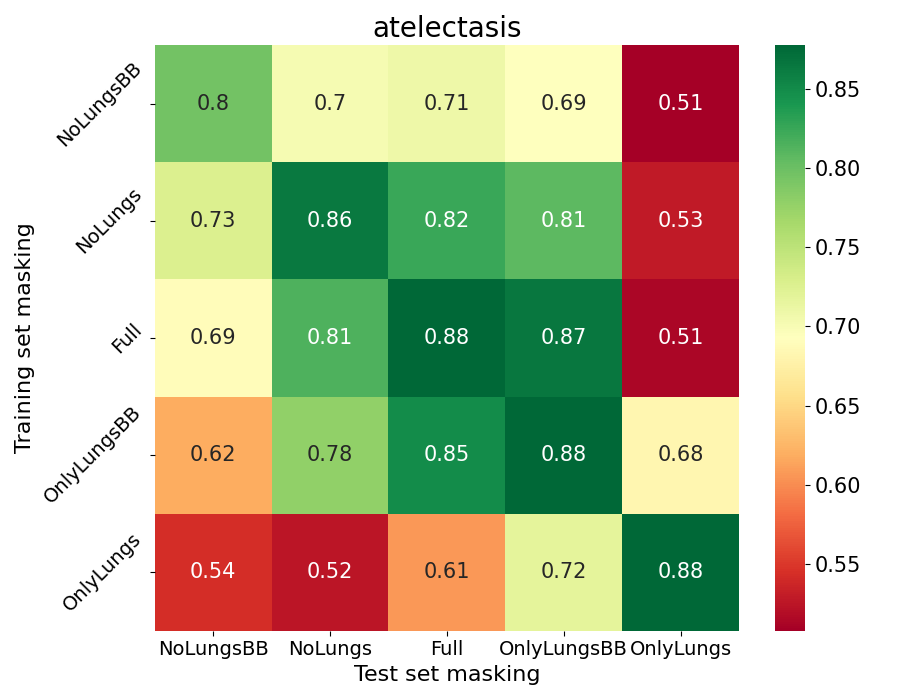}
        \caption{on test set}
    \end{subfigure}
    \caption{Atelectasis mean AUC across 5-fold models with different masking for training and evaluation images}
    \label{fig:mean_auc_atelectasis}
\end{figure}

\begin{figure}[H]
    \centering
    \begin{subfigure}{.45\columnwidth}
        \centering
         \includegraphics[width=\linewidth]{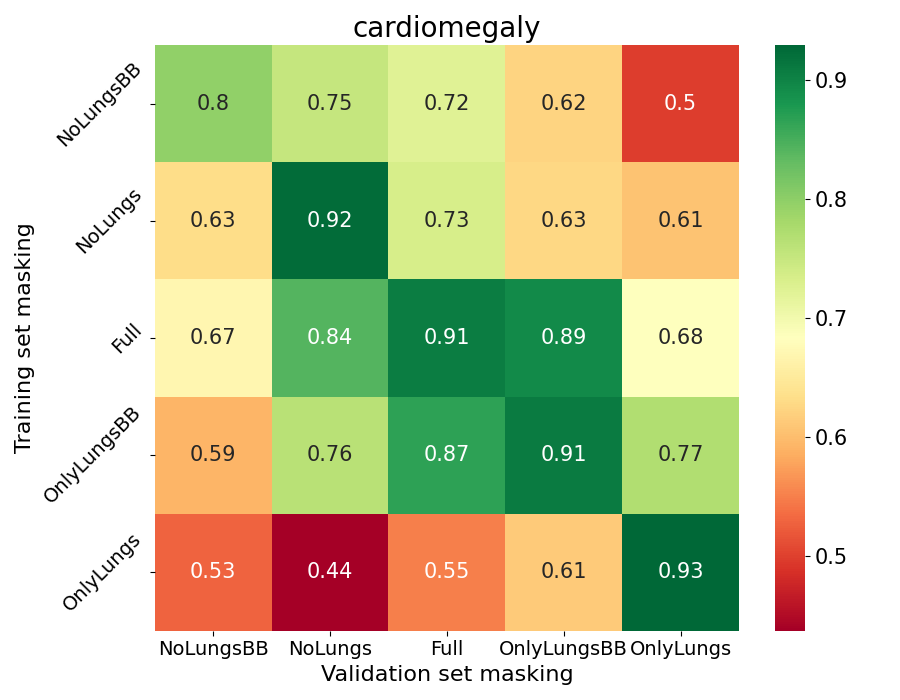}
      \caption{on 5-folds validation sets}     
    \end{subfigure}
    \begin{subfigure}{.45\columnwidth}
      \centering
      \includegraphics[width=\linewidth]{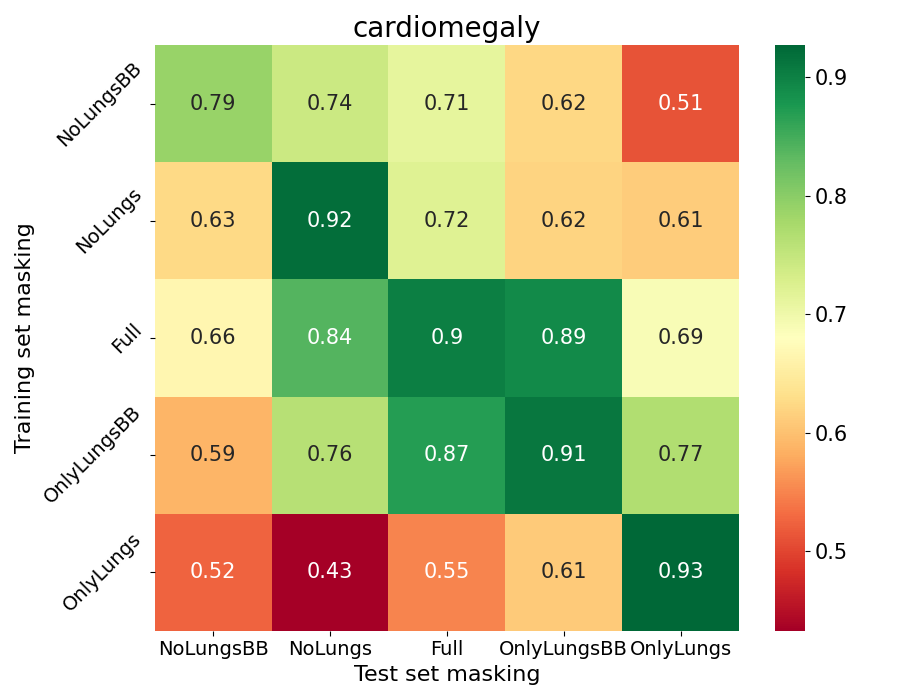}
        \caption{on test set}
    \end{subfigure}
    \caption{Cardiomegaly mean AUC across 5-fold models with different masking for training and evaluation images}
    \label{fig:mean_auc_cardiomegaly}
\end{figure}

\begin{figure}[H]
    \centering
    \begin{subfigure}{.45\columnwidth}
        \centering
         \includegraphics[width=\linewidth]{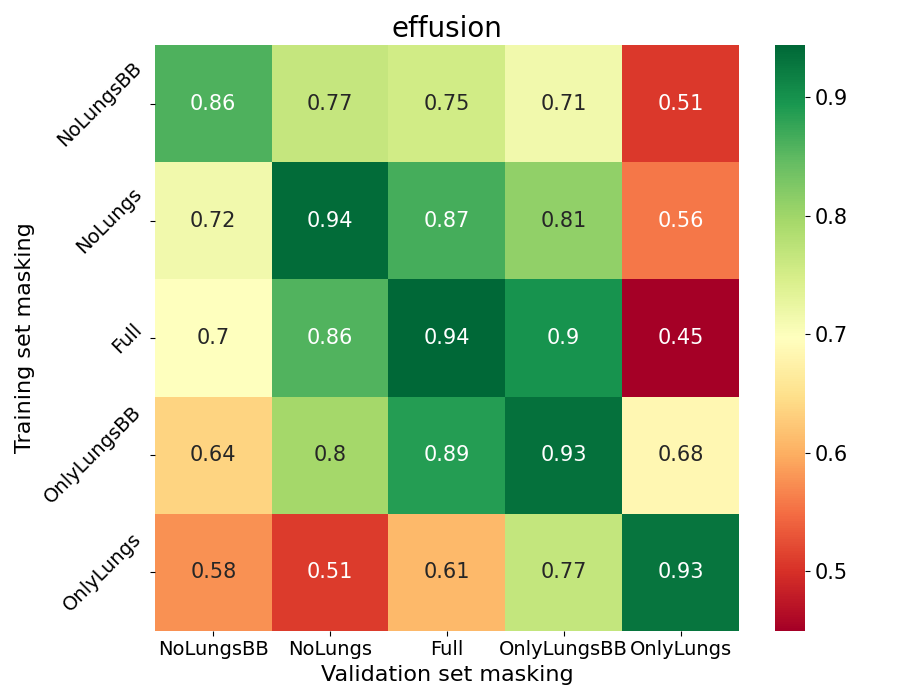}
      \caption{on 5-folds validation sets}     
    \end{subfigure}
    \begin{subfigure}{.45\columnwidth}
      \centering
      \includegraphics[width=\linewidth]{images/test_mean_auc_effusion.png}
        \caption{on test set}
    \end{subfigure}
    \caption{Effusion mean AUC across 5-fold models with different masking for training and evaluation images}
    \label{fig:mean_auc_effusion}
\end{figure}

\begin{figure}[H]
    \centering
    \begin{subfigure}{.45\columnwidth}
        \centering
         \includegraphics[width=\linewidth]{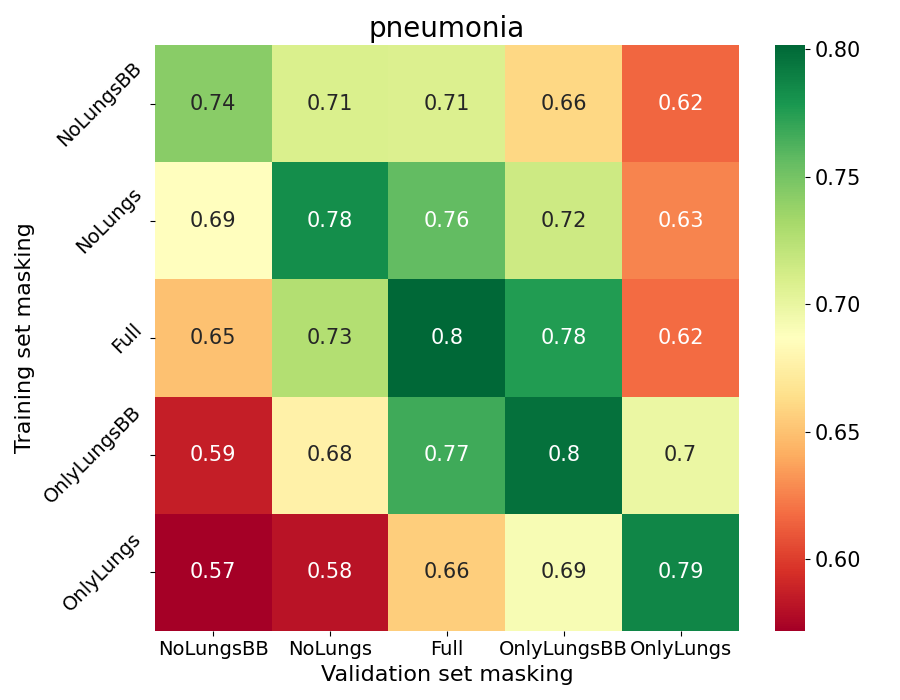}
      \caption{on 5-folds validation sets}     
    \end{subfigure}
    \begin{subfigure}{.45\columnwidth}
      \centering
      \includegraphics[width=\linewidth]{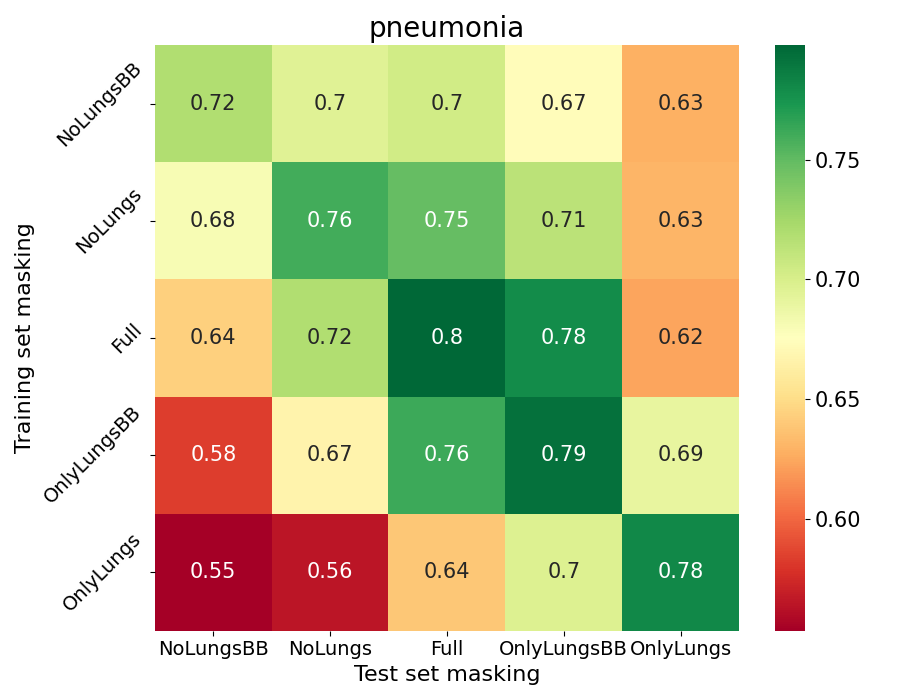}
        \caption{on test set}
    \end{subfigure}
    \caption{Pneumonia mean AUC across 5-fold models with different masking for training and evaluation images}
    \label{fig:mean_auc_pneumonia}
\end{figure}

\begin{figure}[H]
    \centering
    \begin{subfigure}{.45\columnwidth}
        \centering
         \includegraphics[width=\linewidth]{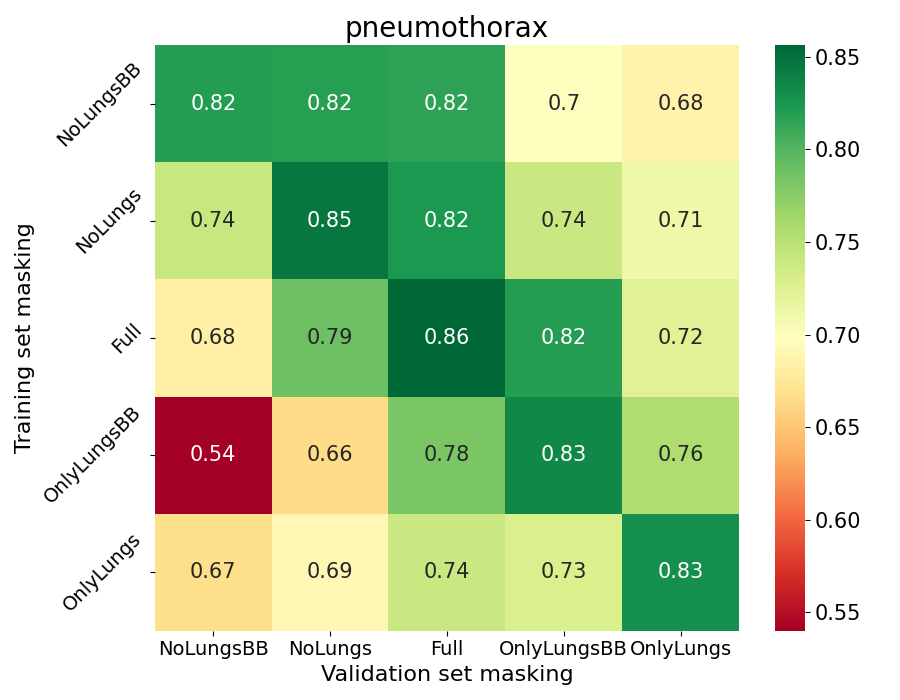}
      \caption{on 5-folds validation sets}     
    \end{subfigure}
    \begin{subfigure}{.45\columnwidth}
      \centering
      \includegraphics[width=\linewidth]{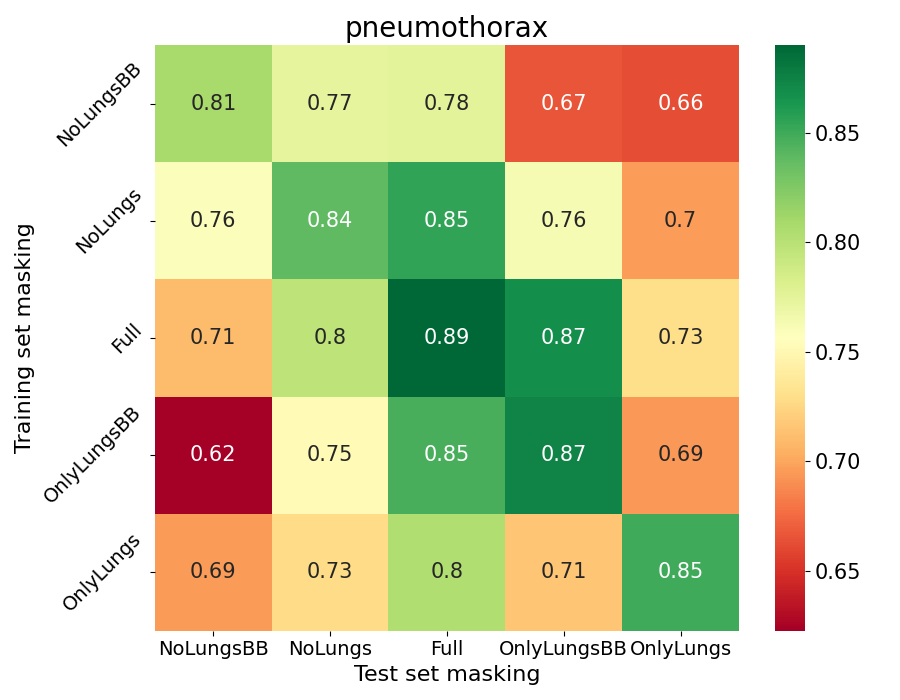}
        \caption{on test set}
    \end{subfigure}
    \caption{Pneumothorax mean AUC across 5-fold models with different masking for training and evaluation images}
    \label{fig:mean_auc_pneumothorax}
\end{figure}

\begin{figure}[H]
    \centering
    \begin{subfigure}{.45\columnwidth}
        \centering
         \includegraphics[width=\linewidth]{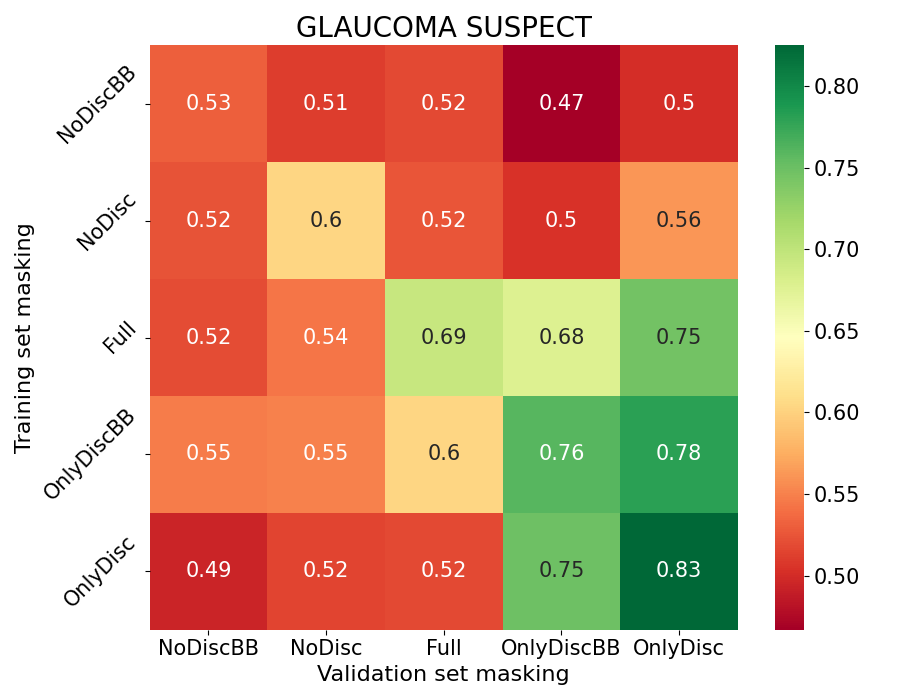}
      \caption{on 5-folds validation sets}     
    \end{subfigure}
    \begin{subfigure}{.45\columnwidth}
      \centering
      \includegraphics[width=\linewidth]{images/test_mean_auc_glaucoma.png}
        \caption{on test set}
    \end{subfigure}
    \caption{Glaucoma mean AUC across 5-fold models with different masking for training and evaluation images}
    \label{fig:mean_auc_glaucoma}
\end{figure}

\section{Example of images with different dilation factors}\label{appendix:dilation}

\begin{figure}[H]
    \centering
    \begin{subfigure}{\columnwidth}
        \centering
         \includegraphics[width=\linewidth]{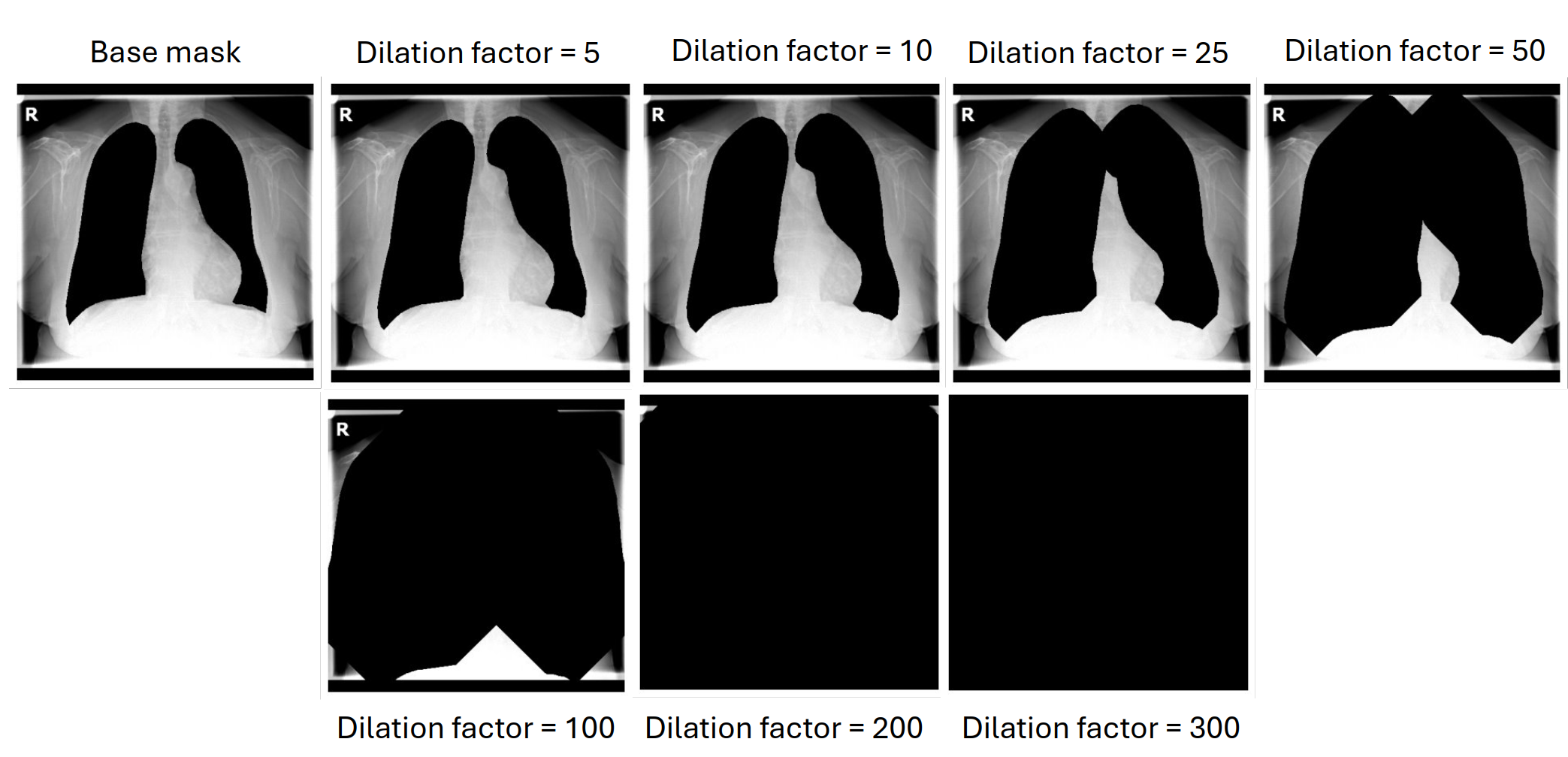}
      \caption{Images without lungs}
      \label{subfig:dilation_NoLungs}
    \end{subfigure}
    \begin{subfigure}{\columnwidth}
      \centering
      \includegraphics[width=\linewidth]{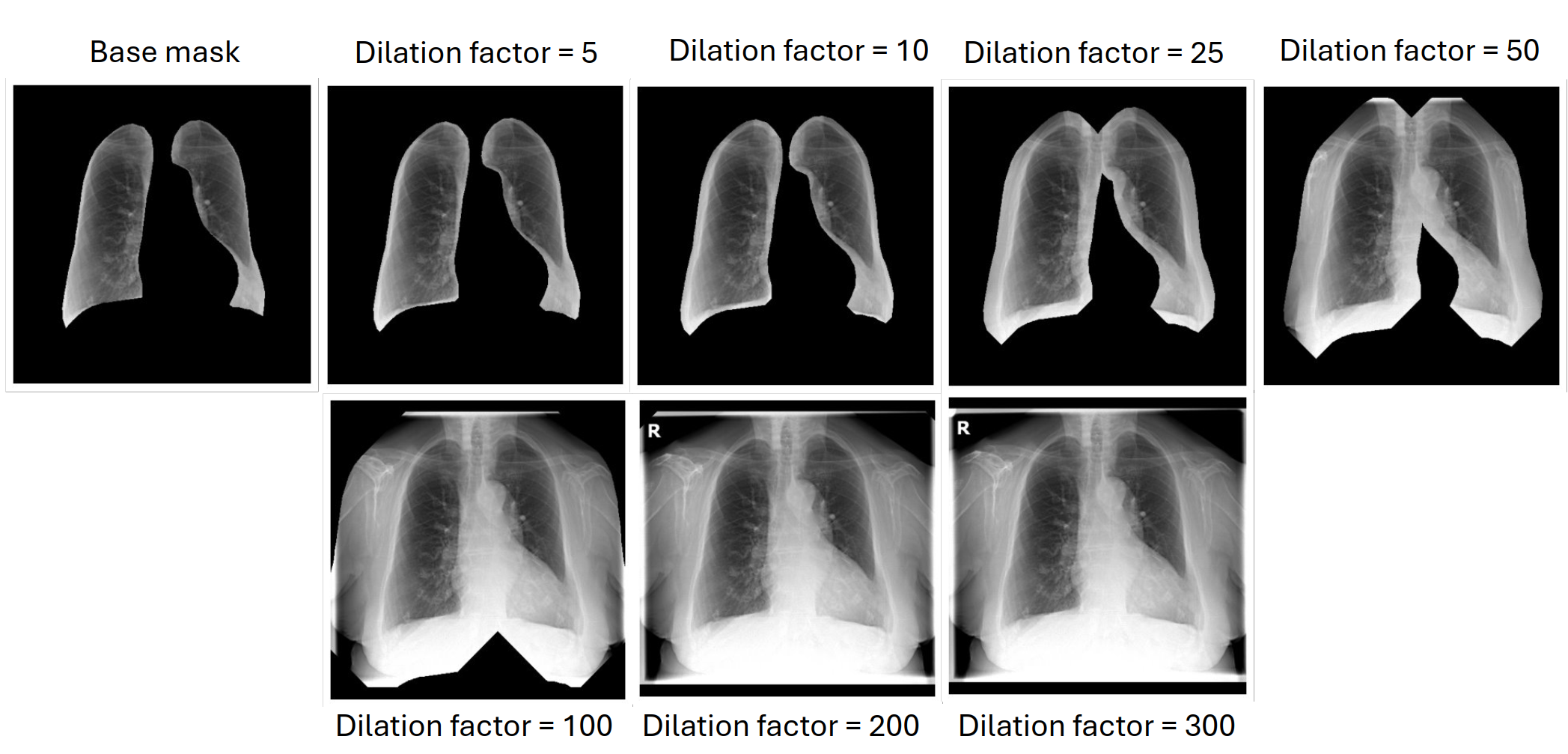}
        \caption{Images with only lungs}
        \label{subfig:dilation_OnlyLungs}
    \end{subfigure}
    \caption{Example of NoLungs and OnlyLungs images while increasing the mask size. In this example, the mask covers the entire image at a dilation factor of 300 or more, but a higher dilation may be needed depending on the original mask.}
    \label{fig:dilation_lungs}
\end{figure}

\begin{figure}[H]
    \centering
    \begin{subfigure}{\columnwidth}
        \centering
         \includegraphics[width=\linewidth]{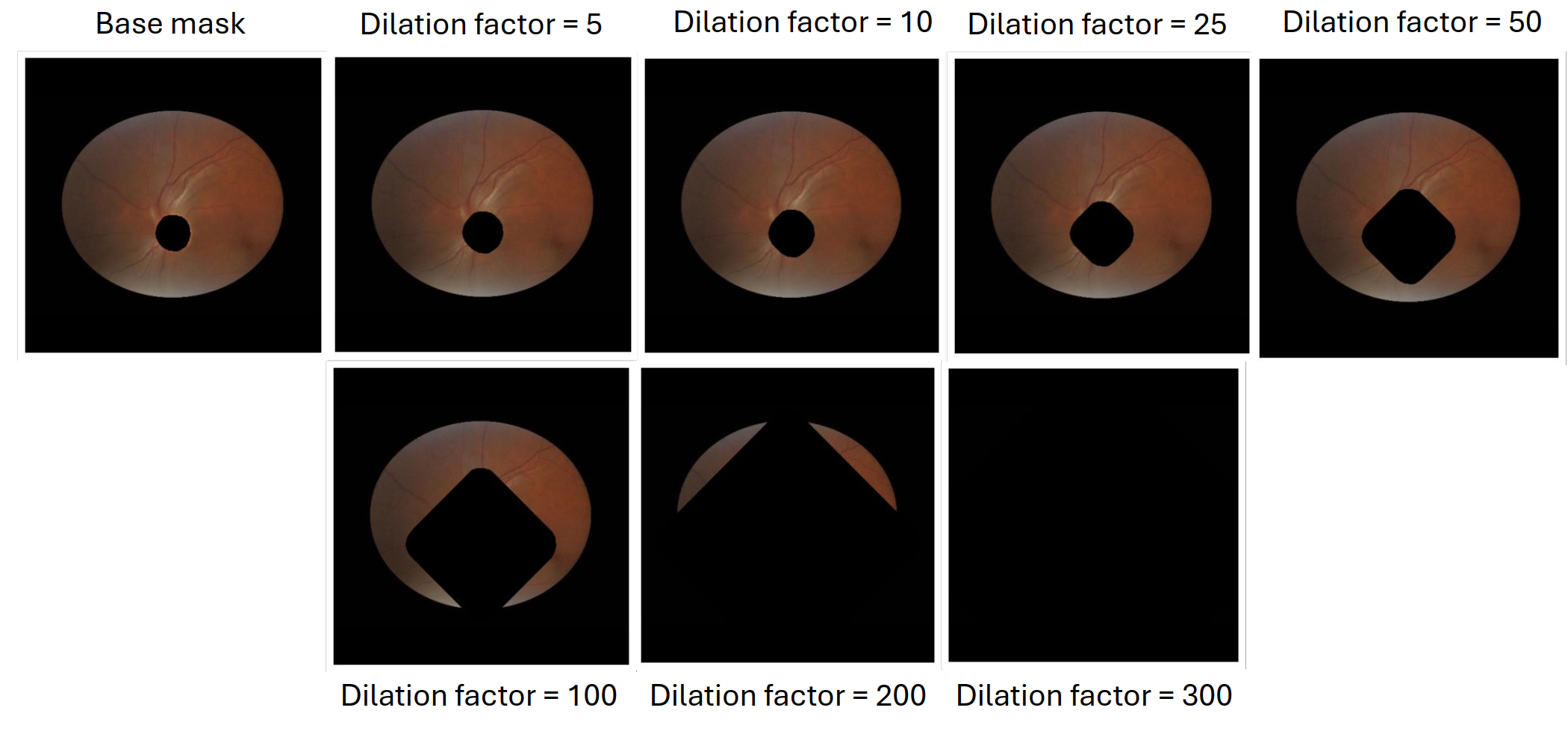}
      \caption{Images without the optic disc}
      \label{subfig:dilation_NoDisc}
    \end{subfigure}
    \begin{subfigure}{\columnwidth}
      \centering
      \includegraphics[width=\linewidth]{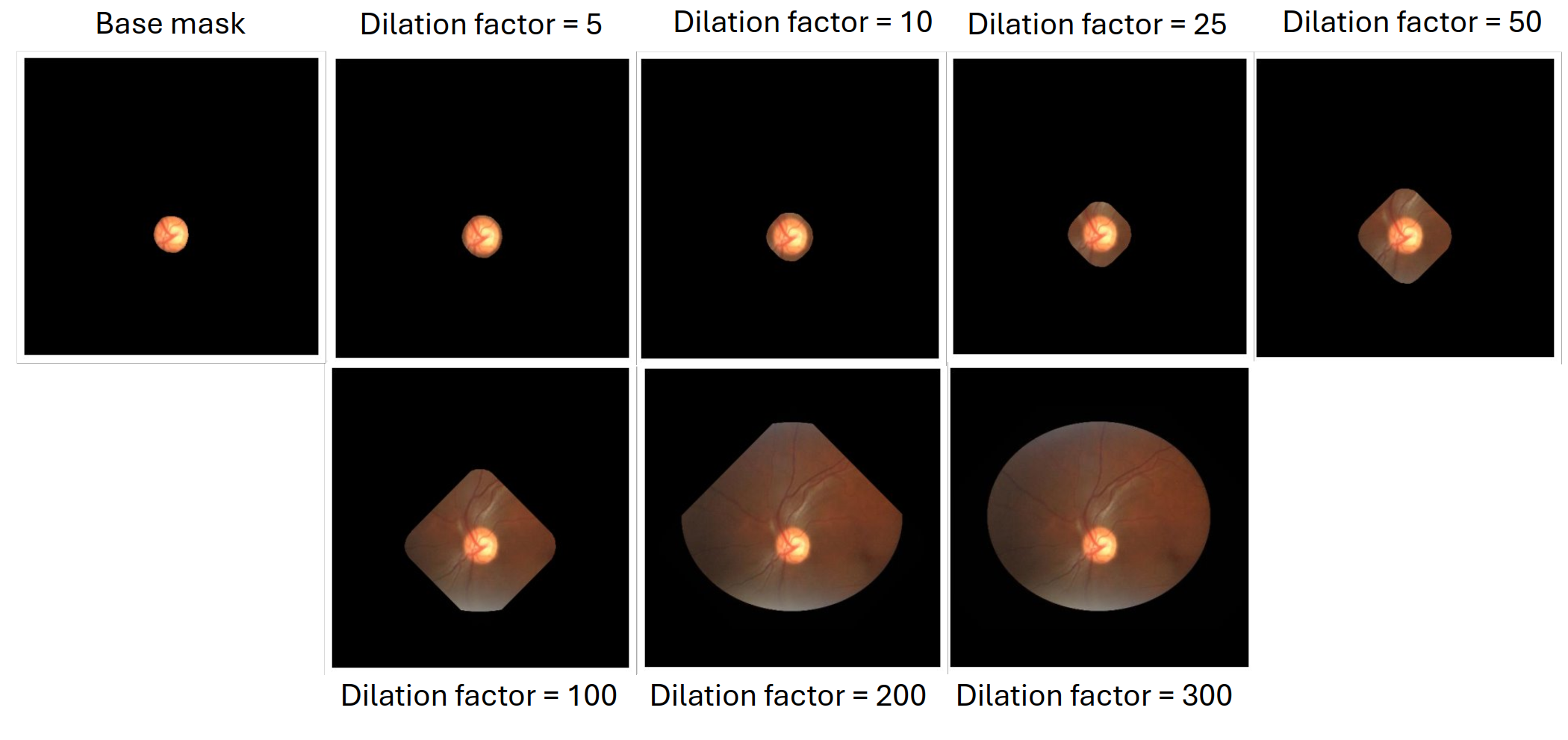}
        \caption{Images with only the optic disc}
        \label{subfig:dilation_OnlyDisc}
    \end{subfigure}
    \caption{Example of NoDisc and OnlyDisc images while increasing the mask size.}
    \label{fig:dilation_eye}
\end{figure}

\end{document}